\documentclass{cambrian} %

\usepackage{amsmath,amsfonts,bm}

\def\eqref#1{equation~\ref{#1}}

\def\1{\bm{1}}

\def\rva{{\mathbf{a}}}

\def\rvc{{\mathbf{c}}}

\def\rvf{{\mathbf{f}}}
\def\rvg{{\mathbf{g}}}
\def\rvh{{\mathbf{h}}}

\def\rvm{{\mathbf{m}}}

\def\rvo{{\mathbf{o}}}
\def\rvp{{\mathbf{p}}}
\def\rvq{{\mathbf{q}}}

\def\rvv{{\mathbf{v}}}

\DeclareMathAlphabet{\mathsfit}{\encodingdefault}{\sfdefault}{m}{sl}
\SetMathAlphabet{\mathsfit}{bold}{\encodingdefault}{\sfdefault}{bx}{n}

\def\gL{{\mathcal{L}}}

\def\gN{{\mathcal{N}}}

\usepackage[utf8]{inputenc} %
\usepackage[T1]{fontenc}    %
\usepackage{natbib}         %
\usepackage{hyperref}
\usepackage{url}
\usepackage{duckuments}
\usepackage{adjustbox}
\usepackage{array}
\usepackage{booktabs}
\usepackage{subcaption}
\usepackage{lipsum}
\usepackage{amsmath}
\usepackage{float}
\usepackage{multirow}
\usepackage{enumitem}
\usepackage{listings}
\usepackage{algorithm}
\usepackage{algpseudocode}
\usepackage{bibunits}
\usepackage{changepage}
\usepackage[capitalize]{cleveref} %
\usepackage{makecell}
\usepackage{wrapfig}
\usepackage{xspace}
\usepackage{pgfplots}

\newcommand{\worldwideweb}{\raisebox{-1.5pt}{\includegraphics[height=1.05em]{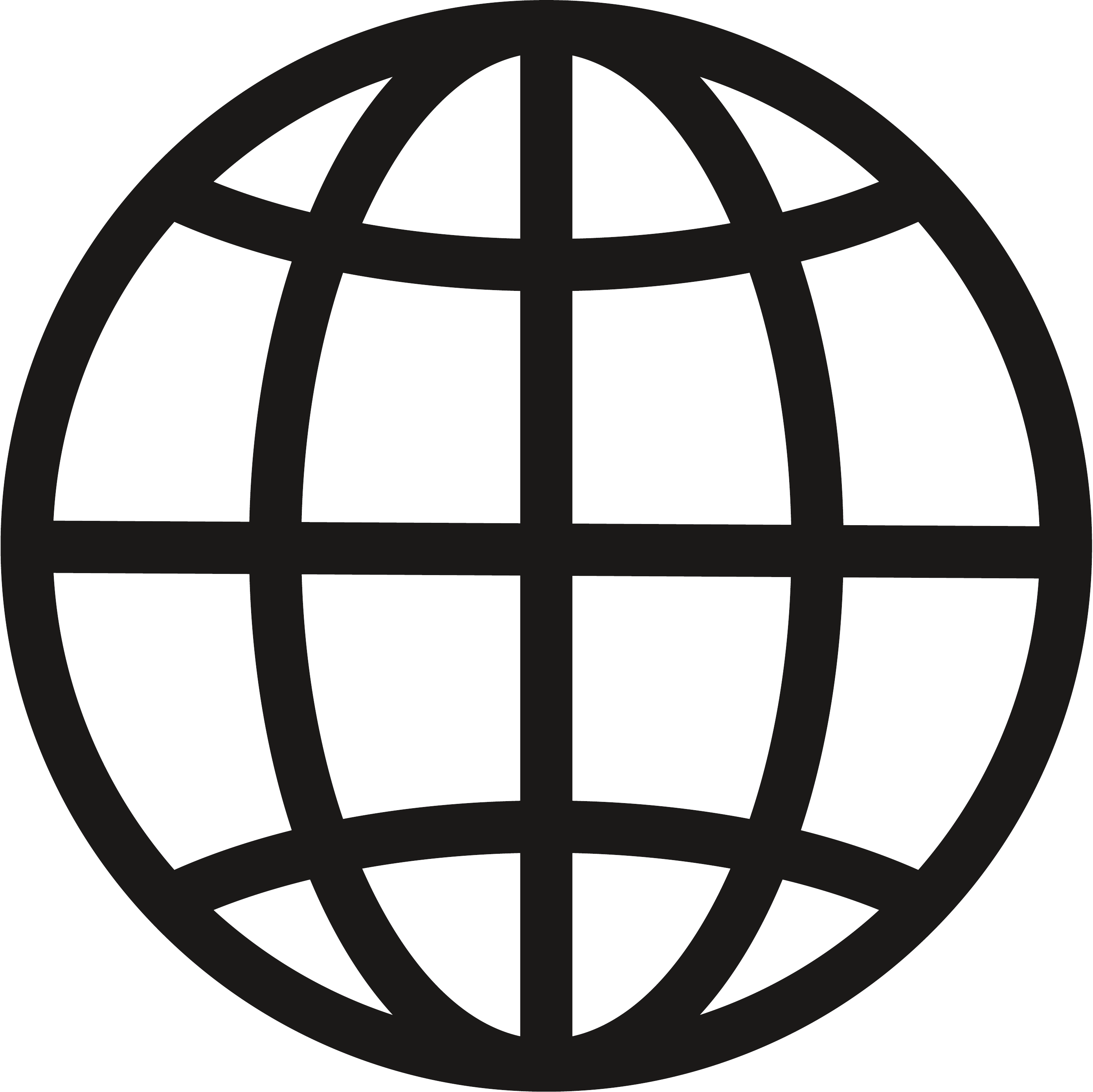}}\xspace}
\newcommand{\huggingface}{\raisebox{-1.5pt}{\includegraphics[height=1.05em]{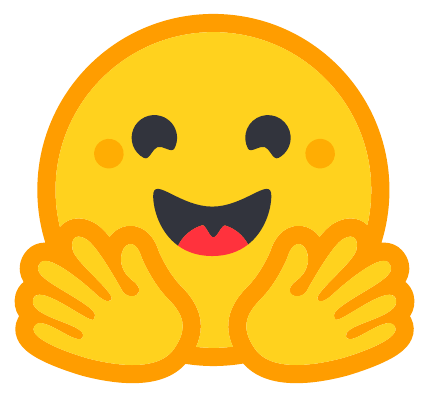}}\xspace}
\newcommand{\github}{\raisebox{-1.5pt}{\includegraphics[height=1.05em]{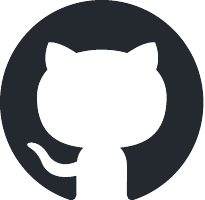}}\xspace}

\usepackage{titlesec}
\titlespacing*{\paragraph}
    {0pt}     %
    {0.25em}  %
    {1em}  %
\definecolor{cornellred}{rgb}{0.7, 0.11, 0.11}
\definecolor{cadmiumgreen}{rgb}{0.0, 0.42, 0.24}
\definecolor{aliceblue}{rgb}{0.91, 0.94, 0.97}
\definecolor{darkblue}{rgb}{0.83, 0.89, 0.97}
\definecolor{Red7}{rgb}{0.941, 0.243, 0.243}
\definecolor{Green7}{RGB}{55, 178, 77}
\definecolor{Blue9}{rgb}{0.098,0.3,0.9}
\definecolor{SJViolet}{RGB}{105,100,171}
\definecolor{SJRed}{RGB}{237,109,107}
\definecolor{baselinecolor}{HTML}{EEEEEE}
\newcommand{\baseline}[1]{\cellcolor{baselinecolor}{#1}}

\usepackage{pifont}
\newcommand{\ie}{\textit{i}.\textit{e}., }
\newcommand{\eg}{\textit{e}.\textit{g}., }

\newcommand{\placeholder}[1]{{\color{lightgray}\lipsum[1]}}
\newcommand{\cmark}{\ding{51}}%
\newcommand{\xmark}{\ding{55}}%

\newcommand{\sname}{FRED\xspace}

\hypersetup{
  linkcolor = cornellred,
  citecolor  = cadmiumgreen,
  colorlinks = true,
  urlcolor = Blue9
}

\pgfplotsset{compat=1.18}

\title{\center{Self-Distilling Future Video Representations \\ for World Action Models}}

\renewcommand\Affilfont{\normalfont\fontsize{11}{15}\selectfont\centering}
\author{
    Huiwon Jang$^{1,2*}$\,
    Suhyeok Jang$^{1,2*}$\,
    Hyeonmo Kang$^{1}$\,
    Seungku Kim$^{1}$\,
    Jinwoo Shin$^{1,2}$\,
    Sihyun Yu$^{1}$\\
    \vspace{0.03in}
    $^{1}$KAIST\quad
    $^{2}$RLWRLD
}

\begingroup

\footnotetext{$^*$ Equal contribution.}
\endgroup
\begin{abstract}
World Action Models (WAMs) have emerged as a promising approach to learning generalist robot policies using rich spatiotemporal priors from pre-trained video diffusion models.
A recent line of WAMs improves inference efficiency by conditioning a separate action decoder on intermediate future video representations extracted from a single denoising step, avoiding costly iterative denoising required for video generation.
This design makes action prediction depend directly on the quality of these future representations.
In this paper, we introduce {\textbf{F}uture \textbf{RE}presentation learning via self-\textbf{D}istillation with temporal context} (\sname), a framework that trains WAMs to improve the quality of future video representations from noise in a single denoising step.
Specifically, \sname uses self-distillation to predict future representations from noise, using representations extracted from ground-truth future observations as targets.
We further condition the video model on historical observations, providing temporal context for predicting future representations better.
Across diverse real-world manipulation scenarios, \sname consistently outperforms competing WAMs. It achieves particularly large gains on long-horizon and out-of-domain tasks under zero-shot environment generalization, and substantially improves human-to-robot skill transfer, reaching 50.0\% task success on tasks absent from robot training data and demonstrated only in human videos, compared with {10.7}\% for the best competing WAM. 
\end{abstract}

\begin{document}

\maketitle

\begin{center}
    \renewcommand{\arraystretch}{1.5}
    \begin{tabular}{rll}
        \worldwideweb{} & \textbf{Project Page} & \url{https://huiwon-jang.github.io/FRED/}\\
        \github{} & \textbf{Code} & \url{https://github.com/suhyeok-jang/FRED/}\\
        \huggingface{} & \textbf{FRED Models} & \url{https://huggingface.co/FRED-AGI-LAB/models} \\
    \end{tabular}
\end{center}

\newpage
{
    \hypersetup{linkcolor=black}
    \setlength{\parskip}{0pt}
    \tableofcontents
}
\newpage

\section{Introduction}

World Action Models (WAMs; \citealt{pai2025mimic,liang2025video,kim2026cosmos,ye2026world}) have recently emerged as a promising approach to learning generalist robot policies. Built on pre-trained video diffusion models (VDMs; \citealt{wan2025wan,ali2025world}), WAMs have shown strong generalization to unseen tasks \citep{ye2026world,ma2026dit4dit}, adaptation to new robot embodiments \citep{ye2026world}, and skill transfer from human videos to robots \citep{dyna2026dyna2}. Their core idea is to leverage the rich spatiotemporal priors learned from large-scale in-the-wild video datasets \citep{chen2024panda,nan2024openvid} by training a model to jointly predict future frames and corresponding actions. This allows the strong generalization capabilities of pre-trained VDMs to extend from future video prediction to robot control.

Recent works have shown that WAMs can achieve competitive performance with a separate action decoder that generates actions conditioned on intermediate hidden states of the VDM \citep{ma2026dit4dit,dyna2026dyna2,qiu2026making,zhao2026faster}. These hidden states are obtained from a single forward pass with Gaussian noise $\bm{\epsilon} \sim \gN(\mathbf{0}, \mathbf{I})$ as input.
This design substantially reduces inference latency by avoiding iterative denoising with large VDM backbones, making WAMs more practical for robot control. However, because the action decoder directly leverages these hidden states, policy performance depends on how well they capture future robot-object dynamics.

\begin{figure*}[t]
\vspace{-0.30in}
\centering\small
\begin{subfigure}[t]{0.4\textwidth}
  \centering
  \includegraphics[width=0.92\linewidth]{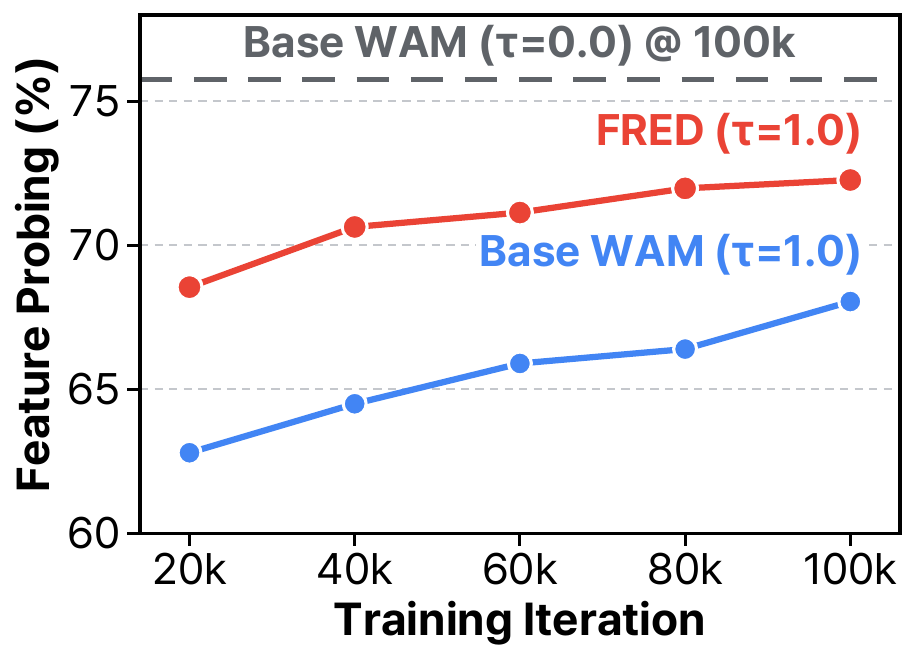}
  \subcaption{Future-motion probing accuracy of \\ VDM hidden states}
  \label{fig:motivation:probing}
\end{subfigure}~
\begin{subfigure}[t]{0.4\textwidth}
  \centering
  \includegraphics[width=0.92\linewidth]{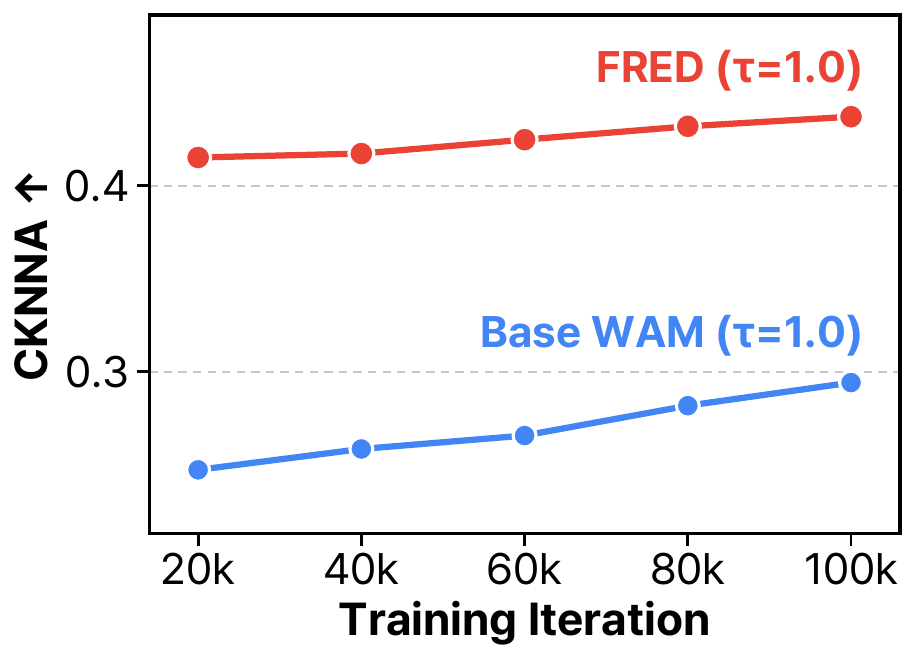}
  \subcaption{Structural similarity to representations extracted from clean future observations}
  \label{fig:motivation:cknna}
\end{subfigure}
\vspace{-0.1in}
\caption{
\textbf{VDM learns future representations during WAM training, and \sname further improves them.} With DiT4DiT as the base WAM, (a) future-motion probing accuracy of hidden states generated from noise (\ie $\tau=1.0$) increases with training but remains below that of representations extracted from clean future observations (\ie $\tau=0.0$); and (b) their structural similarity to representations extracted from clean future observations, measured by CKNNA \citep{huh2024platonic}, also increases during training. The evaluation setup is provided in \Cref{appendix:analysis_setup}.
}
\label{fig:motivation}
\vspace{-0.22in}
\end{figure*}

In this paper, we aim to improve the control performance of the recent line of WAMs that use a single VDM forward pass for action generation.
We start by performing an analysis of how well the VDM hidden states learn meaningful information to generate actions during WAM training.
We first find that VDM hidden states generated from noise already capture information about future robot motion. Specifically, a lightweight probe trained on frozen hidden states can predict future motion primitives, with classification accuracy increasing throughout WAM training (see \Cref{fig:motivation:probing}).
Moreover, these hidden states become increasingly similar to those extracted from ground-truth future observations at the same VDM layer (see \Cref{fig:motivation:cknna}).
These findings suggest that the VDM progressively learns to generate future video representations from noise. 
However, a clear gap in future-motion probing accuracy remains relative to representations extracted from clean future observations, suggesting that generating sufficiently informative future representations from noise remains challenging.

\begin{figure*}[t]
\vspace{-0.30in}
    \centering    \includegraphics[width=.9\textwidth]{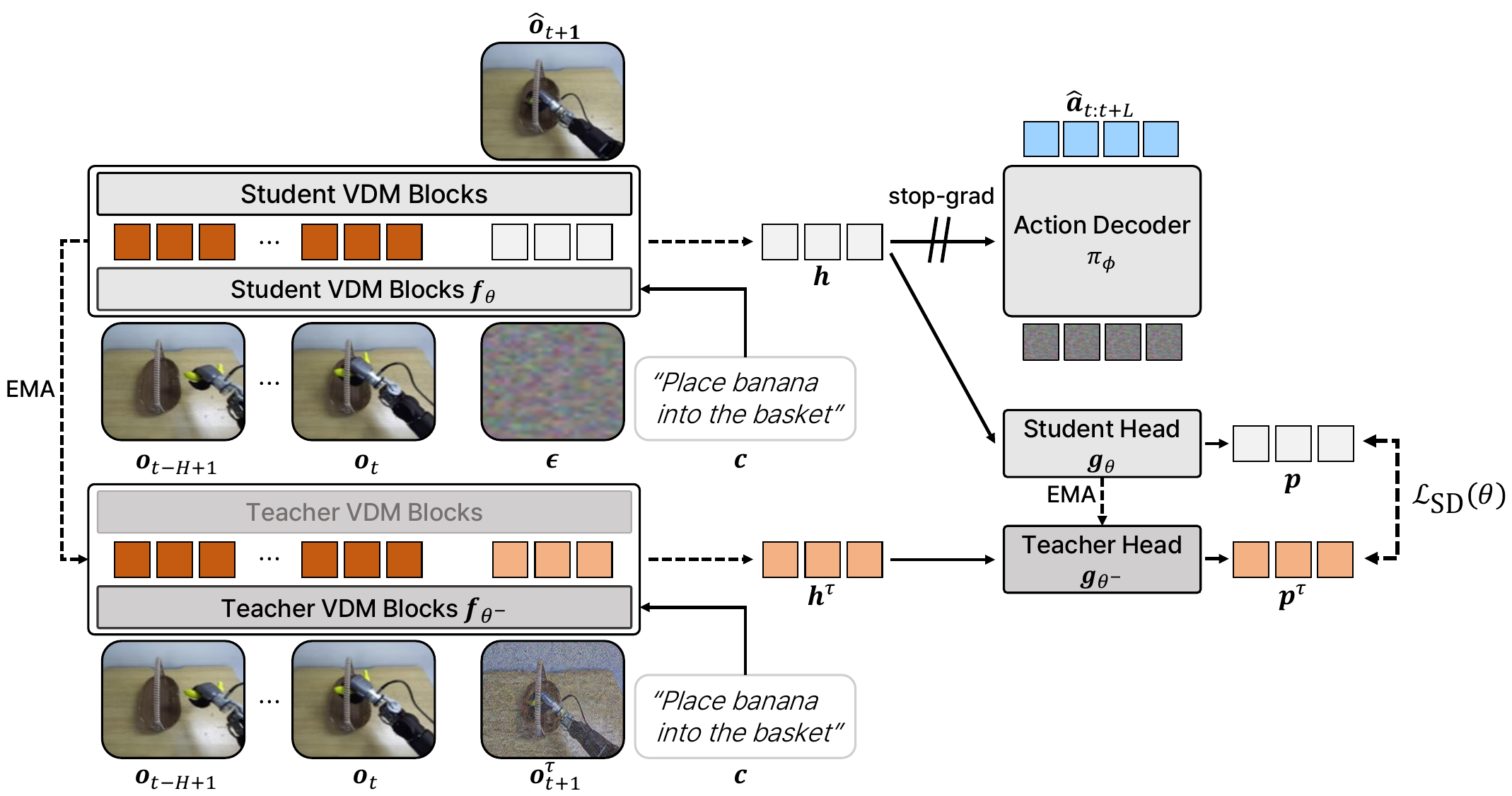}
    \caption{\textbf{Overview of \sname.} 
    We interpret the usage of a VDM in WAMs as a future representation generator, \ie $\rvh = \rvf_\theta(\bm{\epsilon}|\rvo_t, \rvc)$. From this perspective, we enhance $\rvh$ through self-distillation to predict target representations $\rvh^\tau$ extracted from noisy future observations $\rvo_{t+1}^\tau$ by a momentum teacher, while using historical observations $\rvo_{t-H:t}$ as conditioning input to exploit richer temporal context.}
\label{fig:method_overview}
\vspace{-0.1in}
\end{figure*}

Inspired by this, we propose {\textbf{F}uture \textbf{RE}presentation learning via self-\textbf{D}istillation with temporal context} (\sname), a framework for training WAMs to improve the future representations obtained from a single video denoising step for action generation; see the overview in \Cref{fig:method_overview}.
We train the VDM through self-distillation to align representations generated from noise with those extracted from ground-truth future observations, encouraging the model to capture richer future information for predicting actions.
Specifically, \sname uses a momentum teacher, implemented as an Exponential-Moving-Average (EMA) copy of the online VDM, to extract the target representations from noisy future observations at the same intermediate layer used for action conditioning.
We further condition the VDM on a short history of observations, providing temporal context that helps the intermediate hidden states better represent future robot motion. 
As a result, \sname reduces the future-motion probing performance gap to representations extracted from clean future observations and further improves structural similarity to them (see \Cref{fig:motivation}). Notably, this closer alignment with representations of future observations translates into better robot control performance (see \Cref{fig:observation:sr_repgen}).

To validate the generality and effectiveness of \sname, we conduct extensive experiments in diverse real-world robotic manipulation scenarios.
We first evaluate \sname on zero-shot environment generalization using a varied set of real-world manipulation tasks, including long-horizon manipulation, out-of-domain tasks, and object grounding.
Our results show that \sname consistently outperforms competing WAMs in these settings, improving long-horizon task progress from 61.7\% to 81.7\% and out-of-domain task success from 11.1\% to 38.9\% at the 14B scale.
We further evaluate \sname on human-to-robot skill transfer by co-training the video model on human videos without action labels, including tasks absent from robot training data.
We show that \sname significantly outperforms competing WAMs, improving task success from 10.7\% to 50.0\%, while achieving an instruction-following rate of 96.4\%.
These results highlight that \sname improves future representation quality and achieves strong robot control in both zero-shot generalization and human-to-robot transfer.

We summarize the main contributions of this paper below:
\begin{itemize}[leftmargin=*,topsep=0.0em, itemsep=0.0em]
\item We interpret the VDM in WAMs as a future representation generator that produces intermediate hidden states containing predictive information about future robot motion in a single forward pass.
\item We present \sname, a framework that improves future video representations generated from noise for action prediction through self-distillation with target representations extracted from ground-truth future observations, and short-history conditioning for richer temporal context.
\item We show that \sname consistently outperforms baseline WAMs across diverse real-world manipulation settings: zero-shot environment generalization and human-to-robot skill transfer.
\end{itemize}

\vspace{-0.15in}
\section{Related Work}
\textbf{World Action Models.}
Prior work studies robot control via World Action Models (WAMs), which build on pre-trained video diffusion models (VDMs) and jointly model future observations and actions using robot data, allowing spatiotemporal priors from large-scale video to guide action generation~\citep{shen2026videovla,li2026causal,kim2026cosmos, ye2026world}. These models generate plausible future visual states for action prediction, but iterative denoising with large video backbones incurs substantial computational cost at inference time. Recent WAMs address this bottleneck by decoupling video and action generation and reducing video denoising to a single step. Within this shared design, some works predict actions without explicitly generating future videos at inference by masking access to future video features~\citep{yuan2026fast} or directly decoding actions from video-backbone states~\citep{li2026light}.
In contrast, another line of work conditions action prediction on future video representations obtained from noise in a single video denoising step~\citep{pai2025mimic, ma2026dit4dit, qiu2026making, zhao2026faster,wang2026openwam}. 
Our work follows the latter design and focuses on how to train WAMs to generate more informative future representations for action generation.

\textbf{Diffusion Models and Representation Learning.}
Prior works have shown that diffusion denoisers learn discriminative representations useful for visual recognition~\citep{xiang2023denoising,yang2023diffusion,li2023dreamteacher}. More recent work improves these representations during generative training, where REPA~\citep{yu2024representation} aligns noisy diffusion features with clean image
representations from a pre-trained visual encoder, while SRA~\citep{jiang2025no} uses lower-noise representations from the diffusion model itself as targets.
This direction has been extended to VDMs through alignment with pre-trained video representations~\citep{zhang2026videorepa,lee2025improving,bhowmik2026moalign}, while Self-Flow~\citep{chefer2026self} incorporates self-supervised representation learning directly into flow matching without an external encoder.
Our work studies representation learning in WAMs, where intermediate video representations generated from a single denoising step directly condition action generation.
AGRA~\citep{qiu2026making} also adapts representation alignment to this setting, but it aligns these representations with features from a frozen visual encoder.
In contrast, \sname uses representations extracted from observed future frames by VDM itself as distillation targets for representations generated from noise.
This improves the future video representation used by the action decoder without aligning the intermediate video features to a fixed external representation.

\section{Method}

\subsection{Overview}\label{sec:method:overview}

Let $\mathbf{X}=[(\rvo_t,\rvq_t,\rva_t)_{t=1}^{T}, \rvc]$ be an expert demonstration consisting of a visual observation $\rvo_t$, proprioceptive state $\rvq_t$, and robot action $\rva_t$ at each timestep $t \in [1, T]$, with a task instruction $\rvc$. Given expert demonstrations $\mathbf{X}$ from a dataset, we train a policy built on a World Action Model (WAM; \citealt{ye2026world}) which repurposes a pre-trained video diffusion model (VDM) $\rvv_\theta(\rvo_{t+1}^{\tau_v} \mid \rvo_t, \rvc)$ for joint video-action prediction with diffusion timestep $\tau_v$.
Here, the VDM $\rvv_\theta$ predicts future frames $\rvo_{t+1}$ from the current frame $\rvo_t$ and a text instruction $\rvc$ through the iterative denoising process at timestep $\tau_v$.
In particular, our method is motivated by recent progress in the WAM literature \citep{ma2026dit4dit,mimic2026fluxmimic,qiu2026making}.
They keep the VDM architecture unchanged and use a separate diffusion-based decoder to denoise the future action, \ie $\pi_\phi(\rva_t^{\tau_a}|\rvq_t,\rvh)$\footnote{For simplicity, we describe our method as a policy that predicts a single action and a single future observation. In practice, it predicts $L>1$ future actions and observations at once, which is straightforward to implement within the WAM training framework.} with the diffusion timestep $\tau_a$, where $\rvh$ is an intermediate hidden state of  $\rvv_\theta$ obtained at $\tau_v=1$, \ie through a single forward pass from \emph{Gaussian noise} $\bm{\epsilon} \sim \gN(\mathbf{0}, \mathbf{I})$.

Beyond computational efficiency, this decoupled architecture also offers a useful perspective on the role of the VDM. We interpret its usage as a \emph{future representation generator}: $\rvh = \rvf_{\theta}(\bm{\epsilon} | \rvo_t, \rvc)$, where $\rvf_\theta$ denotes the first layers of $\rvv_{\theta}$ that output $\rvh$. From this perspective, we introduce a framework for training WAMs that improves action generation quality significantly by training the VDM to improve $\rvh$ via self-distillation. 
We illustrate an overview of our method, \sname, in \Cref{fig:method_overview}.

\subsection{\sname} \label{sec:method:our-method}

\sname{} introduces a new framework for training WAMs by improving the representation generator $\rvf_\theta$ to generate more informative future video representations $\rvh$ from noise $\bm{\epsilon}$.
We first use self-distillation to train the student $\rvf_\theta$ to predict target future representations from noise, where the targets are extracted by a momentum teacher $\rvf_{\theta^{-}}$ from noisy ground-truth future observations.
Formally, given the current observation $\rvo_{t}$ and text instruction $\rvc$ as conditioning inputs, we first process random noise $\bm{\epsilon}$ through the student (online) VDM encoder $\rvf_\theta$ to generate intermediate video representations $\rvh$. Under the same conditioning, we process noisy future observations $\rvo_{t+1}^\tau := (1-\tau)\rvo_{t+1} + \tau \bm{\epsilon}$ with $\tau \in [0, 1]$ through the teacher (momentum) VDM encoder $\rvf_{\theta^{-}}$ to obtain target representations $\rvh^{\tau}$:
\begin{equation}
\rvh = \rvf_\theta(\bm{\epsilon} | \rvo_{t}, \rvc)
\quad \text{and} \quad
\rvh^{\tau} = \rvf_{\theta^{-}} (\rvo^{\tau}_{t+1} | \rvo_{t}, \rvc)
\label{eq:feature_extraction}
\end{equation}
We then process these intermediate representations through the patch-wise student and teacher projection heads $\rvg_\theta$, $\rvg_{\theta^{-}}$ and apply softmax functions to obtain distributions $\rvp$, $\rvp^{\tau}$, respectively.
Specifically, we apply softmax with temperature $\gamma_1>0$ to obtain the student distribution $\rvp$. For the teacher distribution $\rvp^{\tau}$, we center the logits by subtracting the mean teacher logits $\rvm$ and apply softmax with a sharpening temperature $\gamma_2>0$ to prevent representation collapse, following \citet{caron2021emerging} and \citet{zhou2021ibot}.
\begin{equation}
\rvp = \mathrm{softmax}\left({
\rvg_\theta(\rvh)}/{\gamma_1}
\right)
\quad \text{and} \quad
\rvp^{\tau} = \mathrm{softmax}\left({(
\rvg_{\theta^{-}}(\rvh^{\tau})-\rvm)} / {\gamma_2}
\right)
\end{equation}
Then, \sname trains $\rvf_\theta$ to generate better future video representations by minimizing the KL divergence between the output distributions $\rvp$ and $\rvp^\tau$, averaged over all $N>0$ patches:
\begin{equation} \gL_{\mathrm{SD}}(\theta) =
\frac{1}{N}
\sum_{i=1}^{N}
\mathrm{KL}\left(
\rvp^\tau_{i} \| \rvp_{i}
\right)
\label{eq}
\end{equation}

\textbf{Historical Context for the VDM Encoders.} To facilitate more effective future representation generation, we additionally condition both the student $\rvf_\theta$ and teacher $\rvf_{\theta^{-}}$ on an $H$-frame short history of observations $\rvo_{t-H:t}:= [\rvo_{t-H+1}, \cdots, \rvo_t]$, rather than only the current frame $\rvo_t$. This provides temporal cues (\textit{e.g.}, recent robot motion) that encourage the VDM to infer how the robot has been moving and better predict the future dynamics.

\textbf{WAM Training Objective and Inference.}
The overall WAM training objective combines the self-distillation loss with the video and action flow-matching losses~\citep{lipman2022flow}, as follows:
\begin{equation} \gL_{\mathrm{video}}(\theta) = 
\gL_{\mathrm{FM}}(\theta)
+
\lambda_{\mathrm{SD}}\gL_{\mathrm{SD}}(\theta)
\quad \text{and} \quad
\gL_{\mathrm{action}}(\phi) = \gL_{\mathrm{FM}}(\phi),
\end{equation}
where $\lambda_{\mathrm{SD}}>0$ is a hyperparameter that balances the flow-matching and self-distillation objectives.
Here, we do not update $\theta$ with the action flow-matching loss $\gL_{\mathrm{FM}}(\phi)$; thus, the representation generator $\rvf_\theta$ is trained only by the video denoising and self-distillation objectives.
We update the parameters of momentum teacher networks as ${\theta^{-}} \leftarrow m\theta^{-} + (1-m)\theta$, where $m\in[0,1]$ is the momentum coefficient given as a hyperparameter.

At inference, we obtain the future representation $\rvh$ from the representation generator $\rvf_\theta$, and then use the action decoder $\pi_\phi$ to generate $\rva_{t}$ conditioned on ($\rvq_t$, $\rvh$) using Euler ODE sampling~\citep{lipman2022flow}.
\section{Experiments}
We design experiments to investigate whether learning to generate useful future video representations using \sname can improve the control performance of World Action Models (WAMs) in complex real-world environments. In particular, we consider the following questions:
\begin{itemize}[leftmargin=*,itemsep=0mm]
\vspace{-0.5em}
\item Can \sname generalize to unseen environments and tasks? (\Cref{tab:droid_real})
\item Can \sname benefit from human video co-training, allowing human-to-robot skill transfer? (\Cref{tab:openarm_real})
\item What is the effect of each component of \sname and how general and efficient is it? (\Cref{tab:method_ablation,tab:analysis_design_choices,tab:backbone_generalization,tab:efficiency_analysis}) 
\item How does future representation quality correlate with control performance? (\Cref{tab:sim_results} and \Cref{fig:observations})
\end{itemize}

\begin{figure}[t]
\vspace{-0.15in}
    \centering    \includegraphics[width=0.95\linewidth]{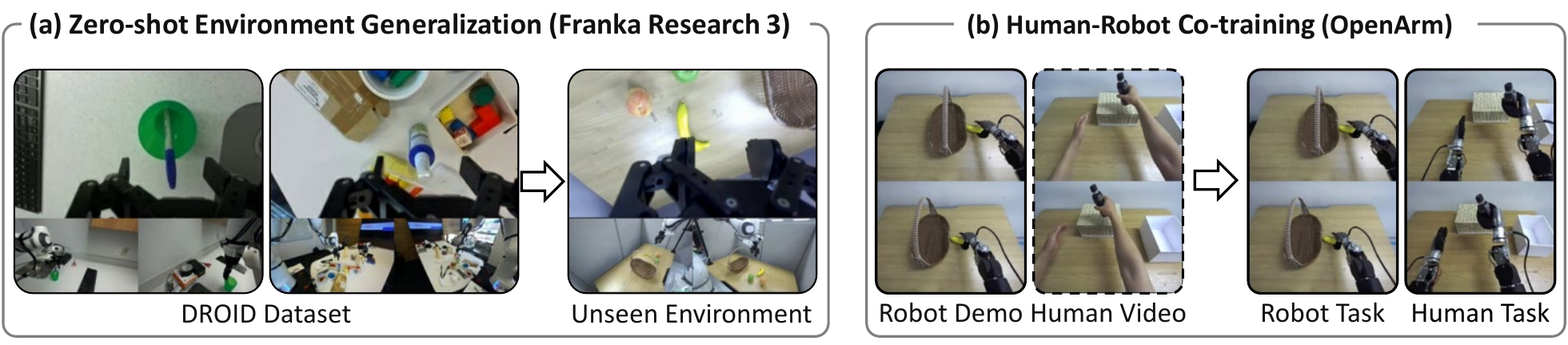}
    \caption{\textbf{Visualization of the training and evaluation setup.} (a) For zero-shot environment generalization, we train \sname on the DROID \citep{khazatsky2024droid} dataset and evaluate it on diverse tasks in new real-world environments without task-specific fine-tuning. (b) For human-robot co-training, we train \sname either on robot demonstrations alone or jointly on robot demonstrations and human videos, and evaluate it on both in-domain robot tasks and human-to-robot skill transfer tasks.}
\end{figure}

\subsection{Experimental Setup}
\label{subsec:experimental_setup}
\textbf{Implementation Details.}
We consider two different model scales for our experiments: 5B and 14B.
For 5B-scale experiments, we use Wan2.2-TI2V-5B~\citep{wan2025wan} as the VDM backbone and a 16-layer DiT~\citep{peebles2023scalable} as the action decoder. For self-distillation in \sname{}, we extract features from the 19th layer (index 18) of the video backbone for both the student and teacher, fix the teacher noise level at 0.25, and use $\lambda_{\mathrm{SD}}=0.01$ by default. We update the momentum teacher using an exponential moving average (EMA) of the student, with the decay warmed up using an inverse decay schedule. We use 4 historical observations with the current observation as input. For 14B-scale experiments, we use Wan2.1-I2V-14B~\citep{wan2025wan} with a 24-layer DiT action decoder and extract features from the 25th layer of the video backbone for self-distillation. Unless otherwise stated, we extract video features from the same layer for self-distillation and action decoder conditioning. See more details in Appendix~\ref{appendix:implementation_details}.

\textbf{Baselines.}
We use several recent robot foundation models as baselines.
Specifically, we first consider recent World Action Models (WAMs): (a) DiT4DiT \citep{ma2026dit4dit}, Fast-WAM \citep{yuan2026fast}, and AGRA \citep{qiu2026making} are WAMs with a single video denoising step, and (b) DreamZero \citep{ye2026world} is a WAM with multi-step video denoising due to shared video-action modeling within the same architecture.
We also consider recent pre-trained Vision-Language-Action models (VLAs): $\pi_{0.5}$ \citep{intelligence2025pi_} and GR00T N1.7 \citep{nvidia2026gr00tn17}. We describe further details in Appendix~\ref{appendix:baselines}.

\subsection{Zero-shot Environment Generalization}
\label{subsec:zero-shot_generalization}
First, we investigate whether \sname can generalize to unseen environments and tasks using future video representations generated in a single denoising step.
To this end, we train our method on DROID \citep{khazatsky2024droid}, one of the largest publicly available robotics datasets, and evaluate it across diverse real-world situations, including unseen environments and tasks.
Specifically, we categorize the evaluation tasks as follows:
\vspace{-0.1in}
\begin{itemize}[leftmargin=*,itemsep=0mm]
\item \textit{In-domain easy tasks} use instruction and object categories seen during training, but are evaluated in new environments with different object details. We evaluate the tasks in two environments with different tabletop appearances.
\item \textit{In-domain long-horizon task} extends in-domain easy settings to longer-horizon manipulations that require multiple manipulation steps.
\item \textit{Out-of-domain tasks} evaluate generalization to task instructions whose action verbs are absent from the DROID training dataset, assessing the ability to map these novel instructions to appropriate manipulation behaviors.
\item \textit{Object grounding tasks} require the robot to select and manipulate the object specified by the instruction in the presence of distractors with similar colors or shapes.
\end{itemize}
\vspace{-0.1in}
We provide more detailed setup in Appendix~\ref{appendix:benchmark_details:droid}, and the full results in Appendix~\ref{appendix:full_results:droid}.

\begin{table*}[t]
\caption{\textbf{Results on zero-shot generalization to new real-world environments.}
Performance of various policies trained on the DROID \citep{khazatsky2024droid} dataset without environment-specific fine-tuning. We report the average task progress rate (\%) for the in-domain long-horizon task and the average success rate (\%) for other task types, with standard errors estimated using stratified bootstrapping \citep{agarwal2021deep}. 
For in-domain easy tasks, Env. 1 and Env. 2 denote two environments with different tabletop appearances. All other tasks are evaluated in Env. 1. $\pi_{0.5}$, GR00T N1.7, and DreamZero use public checkpoints; the remaining 5B and 14B models are trained for 150K and 100K iterations, respectively.
}
\vspace{-0.5em}
\label{tab:droid_real}
\centering\small
\begin{adjustbox}{max width=0.95\textwidth}
\begin{tabular}{l cc cc cc}
\toprule
& 
& \multicolumn{3}{c}{In-domain}
&
& \multirow{3}{*}{\makecell{Object\\Grounding}} \\
\cmidrule(lr){3-5}
Method & Backbone & Easy (Env. 1) & Easy (Env. 2) & Long-horizon &
Out-of-domain & \\
\midrule
$\pi_{0.5}$~\citep{intelligence2025pi_} 
& -- & 55.6{\scriptsize $\pm$5.4} & 47.2{\scriptsize $\pm$6.9} & 53.3{\scriptsize $\pm$9.8} &  \phantom{0}5.6{\scriptsize $\pm$3.8} & 16.7{\scriptsize $\pm$7.4\phantom{0}} \\
GR00T N1.7~\citep{nvidia2026gr00tn17} 
& -- & 41.7{\scriptsize $\pm$5.6} & 41.7{\scriptsize $\pm$4.5} & 60.0{\scriptsize $\pm$6.7} &  \phantom{0}2.8{\scriptsize $\pm$2.7} & 20.8{\scriptsize $\pm$8.2\phantom{0}} \\
\midrule
Fast-WAM \citep{yuan2026fast} & Wan2.2-5B
& 25.0{\scriptsize $\pm$6.6} & 30.6{\scriptsize $\pm$6.8} & 26.7{\scriptsize $\pm$4.3} & \phantom{0}0.0{\scriptsize $\pm$0.0} & \phantom{0}8.3{\scriptsize $\pm$5.6\phantom{0}} \\
DiT4DiT \citep{ma2026dit4dit} & Wan2.2-5B 
& 16.7{\scriptsize $\pm$5.9} & \phantom{0}8.3{\scriptsize $\pm$4.5} & 16.7{\scriptsize $\pm$4.6} & \phantom{0}0.0{\scriptsize $\pm$0.0} & \phantom{0}8.3{\scriptsize $\pm$5.6\phantom{0}} \\
AGRA \citep{qiu2026making} & Wan2.2-5B
& \textbf{41.7}{\scriptsize $\pm$4.2} & 44.4{\scriptsize $\pm$6.8} & 31.7{\scriptsize $\pm$6.0} & \phantom{0}5.6{\scriptsize $\pm$3.8} & 33.3{\scriptsize $\pm$9.5\phantom{0}} \\
\textbf{\sname{} (Ours)} & Wan2.2-5B
& \textbf{41.7}{\scriptsize $\pm$6.6} & \textbf{52.8}{\scriptsize $\pm$7.8} & \textbf{46.7}{\scriptsize $\pm$9.5} & \textbf{19.4}{\scriptsize $\pm$5.9} & \textbf{37.5}{\scriptsize $\pm$9.8\phantom{0}} \\
\midrule
DreamZero \citep{ye2026world} & Wan2.1-14B
& 63.9{\scriptsize $\pm$7.6} & 58.3{\scriptsize $\pm$7.2} & 61.7{\scriptsize $\pm$9.0} & 11.1{\scriptsize $\pm$5.1} & 16.7{\scriptsize $\pm$7.4\phantom{0}} \\
\textbf{\sname{} (Ours)} & Wan2.1-14B 
& \textbf{77.8}{\scriptsize $\pm$6.7} & \textbf{61.1}{\scriptsize $\pm$7.6} & \textbf{81.7}{\scriptsize $\pm$4.4} & \textbf{38.9}{\scriptsize $\pm$8.0} & \textbf{54.2}{\scriptsize $\pm$10.1} \\
\bottomrule
\end{tabular}
\end{adjustbox}
\vspace{-0.5em}
\end{table*}

\textbf{Results.} As shown in \Cref{tab:droid_real}, \sname consistently outperforms or matches the other WAM baselines across all task types in the 5B video backbone setup.
The improvements are particularly large on long-horizon and out-of-domain tasks: compared with the strongest competing WAM baseline, \sname improves long-horizon task progress from 31.7\% to 46.7\% and out-of-domain task success from 5.6\% to 19.4\%, while also improving object grounding success from 33.3\% to 37.5\%.
Compared with DiT4DiT~\citep{ma2026dit4dit}, which does not explicitly regularize intermediate video representations, \sname achieves substantially higher performance on both long-horizon (46.7\% vs.\ 16.7\%) and out-of-domain tasks (19.4\% vs.\ 0.0\%), suggesting that improving future representations can be beneficial for WAMs in challenging generalization settings.
At the larger 14B scale, \sname further improves across all task types, outperforming both state-of-the-art VLAs and DreamZero~\citep{ye2026world}, which uses the same video backbone. Compared with DreamZero, \sname substantially improves long-horizon task progress (81.7\% vs.\ 61.7\%) and out-of-domain task success (38.9\% vs.\ 11.1\%).
Overall, these results show that our future representation learning remains effective at larger model scales with improved generalization.
\vspace{-0.02in}

\subsection{Human-Robot Co-training}
We next investigate whether \sname can leverage human video to improve robot control, as WAMs with separate video and action modules naturally support joint training on robot and human demonstrations by applying the video generation objective to human videos without requiring action labels.
To this end, we first collect robot demonstrations using the OpenArm humanoid\footnote{https://openarm.dev/} and human manipulation videos in the same environment. 
We then compare policies trained on robot demonstrations alone with those jointly trained on robot demonstrations and human videos.
We categorize the evaluation tasks as follows:

\begin{table}[t]
\vspace{-0.15in}
\centering
\caption{\textbf{Results on humanoid control with and without human-robot co-training.} Performance of policies trained on the OpenArm humanoid using robot demonstrations only or with additional human videos. We report the average success rate (\%) on in-domain atomic tasks, the average task progress rate (\%) on the in-domain composite task, and both the average success rate (\%) and instruction-following rate (\%) on human skill transfer tasks, with standard errors estimated using stratified bootstrapping \citep{agarwal2021deep}. All WAMs use a 5B video backbone.
}
\label{tab:openarm_real}
\vspace{-0.7em}
\begin{subtable}{\linewidth}
\centering
\caption{{Results on robot in-domain tasks.}}
\label{tab:openarm_real_robot}
\scalebox{0.9}{
\begin{tabular}{lcccc}
\toprule
& \multicolumn{2}{c}{{Atomic}}
& \multicolumn{2}{c}{{Composite}} \\
\cmidrule(lr){2-3}
\cmidrule(lr){4-5}
Method
& w/o Human 
& w/ Human
& w/o Human
& w/ Human \\
\midrule
$\pi_{0.5}$ \citep{intelligence2025pi_}
    & 63.9{\scriptsize $\pm$6.2} & --   & 50.0{\scriptsize $\pm$0.0}  & --   \\
GR00T N1.7 \citep{nvidia2026gr00tn17}
    & 91.7{\scriptsize $\pm$4.5} & --   & 45.8{\scriptsize $\pm$4.0}  & --   \\
\midrule
DreamZero \citep{ye2026world}
    & 61.1{\scriptsize $\pm$6.6} & 58.3{\scriptsize $\pm$5.3} & \phantom{0}0.0{\scriptsize $\pm$0.0}  & \phantom{0}0.0{\scriptsize $\pm$0.0}  \\
Fast-WAM \citep{yuan2026fast}
    & 55.6{\scriptsize $\pm$7.4} & 63.9{\scriptsize $\pm$6.5} & \phantom{0}0.0{\scriptsize $\pm$0.0}  & \phantom{0}0.0{\scriptsize $\pm$0.0}  \\
DiT4DiT \citep{ma2026dit4dit}
    & 77.8{\scriptsize $\pm$6.0} & 86.1{\scriptsize $\pm$5.5} & 50.0{\scriptsize $\pm$0.0}  & 41.7{\scriptsize $\pm$5.4}  \\
AGRA \citep{qiu2026making}
    & 80.6{\scriptsize $\pm$6.5} & 75.0{\scriptsize $\pm$6.6} & \phantom{0}8.3{\scriptsize $\pm$5.4}   & 12.5{\scriptsize $\pm$6.3}   \\
\textbf{\sname{} (Ours)}
    & \textbf{91.7}{\scriptsize $\pm$4.2}
    & \textbf{97.2}{\scriptsize $\pm$2.7}
    & \textbf{70.8}{\scriptsize $\pm$9.2}
    & \textbf{75.0}{\scriptsize $\pm$9.3} \\
\bottomrule
\end{tabular}}
\end{subtable}

\vspace{0.5em}

\begin{subtable}{\linewidth}
\centering
\caption{Results on human skill transfer tasks.}
\label{tab:openarm_real_h2r}
\scalebox{0.9}{
\begin{tabular}{lcccc}
\toprule
& \multicolumn{2}{c}{{Task Success}}
& \multicolumn{2}{c}{{Instruction Following}} \\
\cmidrule(lr){2-3}
\cmidrule(lr){4-5}
{Method}
& {w/o Human}
& {w/ Human}
& {w/o Human}
& {w/ Human} \\
\midrule
$\pi_{0.5}$ \citep{intelligence2025pi_}
    & 0.0{\scriptsize $\pm$0.0} & --   & 17.9{\scriptsize $\pm$3.1} & --   \\
GR00T N1.7 \citep{nvidia2026gr00tn17}
    & 9.5{\scriptsize $\pm$2.8} & --   & 66.7{\scriptsize $\pm$1.9} & --   \\
\midrule
DreamZero \citep{ye2026world}
    & 0.0{\scriptsize $\pm$0.0} & \phantom{0}1.2{\scriptsize $\pm$1.1}  & \phantom{0}0.0{\scriptsize $\pm$0.0}  & \phantom{0}9.5{\scriptsize $\pm$2.9} \\
Fast-WAM \citep{yuan2026fast}
    & 0.0{\scriptsize $\pm$0.0} & \phantom{0}0.0{\scriptsize $\pm$0.0}  & \phantom{0}0.0{\scriptsize $\pm$0.0}  & \phantom{0}8.3{\scriptsize $\pm$2.8} \\
DiT4DiT \citep{ma2026dit4dit}
    & 0.0{\scriptsize $\pm$0.0} & 10.7{\scriptsize $\pm$3.1} & \phantom{0}0.0{\scriptsize $\pm$0.0}  & 42.9{\scriptsize $\pm$4.4} \\
AGRA \citep{qiu2026making}
    & 0.0{\scriptsize $\pm$0.0} & \phantom{0}0.0{\scriptsize $\pm$0.0}  & \phantom{0}0.0{\scriptsize $\pm$0.0}  & \phantom{0}0.0{\scriptsize $\pm$0.0} \\
\textbf{\sname{} (Ours)}
    & 0.0{\scriptsize $\pm$0.0}
    & \textbf{50.0}{\scriptsize $\pm$5.1}
    & \phantom{0}0.0{\scriptsize $\pm$0.0}
    & \textbf{96.4}{\scriptsize $\pm$1.9} \\
\bottomrule
\end{tabular}}
\end{subtable}
\end{table}

\begin{itemize}[leftmargin=*,itemsep=0mm]
\vspace{-0.1in}
\item \textit{In-domain atomic tasks} evaluate basic manipulation skills present in the robot training data.
\item \textit{In-domain composite task} requires following an unseen instruction that combines two manipulation skills seen during robot training.
\item \textit{Human skill transfer} covers manipulation tasks involving placements, objects, or action verbs presented only in human demonstrations. We evaluate both task success and instruction-following rates.
\end{itemize}
\vspace{-0.1in}
We provide more details in Appendix~\ref{appendix:benchmark_details:openarm}, and the full results in Appendix~\ref{appendix:full_results:openarm}.

\textbf{In-domain Results.}
\Cref{tab:openarm_real_robot} shows that \sname consistently outperforms WAM baselines in the in-domain setting when trained with only robot data.
In particular, \sname shows a larger margin over the baselines on the composite task, achieving 70.8\% task progress compared with 50.0\% for the strongest baseline. This suggests that the benefit of \sname extends beyond executing individual skills to combining them.
We also observe that human video co-training further improves \sname's success rate on atomic tasks from 91.7\% to 97.2\% and task progress rate on the composite task from 70.8\% to 75.0\%, suggesting that our method helps translate human data co-training into consistent improvements.

\textbf{Human Skill Transfer Results.}
In \Cref{tab:openarm_real_h2r}, we observe that \sname achieves strong human-to-robot skill transfer when co-trained with human videos, while other WAM baselines show limited transfer on tasks demonstrated only in the human-video dataset. For example, DreamZero~\citep{ye2026world} and AGRA~\citep{qiu2026making} achieve instruction-following rates of only 9.5\% and 0.0\%, respectively, with task success rates of 1.2\% and 0.0\%, whereas \sname achieves an instruction-following rate of 96.4\% and a task success rate of 50.0\%. Notably, \sname reliably follows the target skills demonstrated in human videos, whereas most baseline WAMs, except DiT4DiT~\citep{ma2026dit4dit}, often execute behaviors seen in the robot training tasks instead. Without human video co-training, \sname and the WAM baselines achieve near-zero task success. The VLA baselines nevertheless show some instruction-following behavior, possibly due to knowledge acquired during VLA pre-training.
These results highlight the effectiveness of \sname in transferring skills from human videos to robot control, suggesting that future prediction in VDMs provides useful motion information for following human task instructions and that explicitly learning future video representations independently of robot action optimization may further improve human-to-robot transfer.

\subsection{Analysis and Ablation Studies}
We conduct extensive ablation studies and analyses to investigate the component contributions, generality, and inference efficiency of \sname, and the correlation between the quality of generated future representations and robot control performance. We provide an evaluation setup in Appendix~\ref{appendix:benchmark_details:sim} and more analysis in Appendix~\ref{appendix:additional_analysis}.

\textbf{Component Analysis.}
In \Cref{tab:method_ablation}, we evaluate the contribution of self-distillation to future representation learning and the role of historical conditioning in providing temporal context.
We find that both components provide complementary gains across real-world and simulation experiments, as removing either component reduces performance, while removing both leads to further degradation.
For instance, on the DROID in-domain long-horizon (LH) task, removing self-distillation or historical conditioning reduces the task progress rate from 43.3\% to 21.7\% or 28.3\%, respectively, while removing both reduces it to 8.3\%.

\textbf{Self-distillation Design Choices.}
We analyze several design choices of \sname: teacher noise level, feature extraction layer used for both self-distillation and action conditioning, and self-distillation loss coefficient. In \Cref{tab:analysis_design_choices:teacher_noise}, we find that a moderately noisy teacher input with $\tau=0.25$ performs best, outperforming both clean future observations and more heavily corrupted inputs.
We hypothesize that moderate corruption provides more informative target representations for self-distillation, consistent with prior observations that representations from diffusion models can be strongest at intermediate noise levels~\citep{yu2024representation,velez2025image}.
In \Cref{tab:analysis_design_choices:distill_layer}, we find that performance peaks at layer index 18, while using either shallower or deeper layers degrades performance, which aligns with prior findings in WAMs that intermediate video model layers provide more effective representations for action conditioning~\citep{ma2026dit4dit}.
Finally, in \Cref{tab:analysis_design_choices:sd_coeff}, we find that $\lambda_{\mathrm{SD}}=0.01$ performs best among the evaluated coefficients, while a larger coefficient of $0.1$ achieves comparable performance.

\textbf{Comparison with Representation Learning Methods.}
To verify the effectiveness of \sname{} as a representation learning method for WAMs, we compare \sname with recent representation learning approaches for diffusion models, \ie VideoREPA \citep{zhang2026videorepa}, SRA \citep{jiang2025no}, and Self-Flow \citep{chefer2026self}, on RoboCasa Kitchen \citep{nasiriany2024robocasa}, GR-1 Tabletop \citep{bjorck2025gr00t}, and LIBERO \citep{liu2023libero} benchmarks.
In \Cref{tab:sim_results}, VideoREPA, SRA, and Self-Flow generally show modest gains over the base WAM, whereas \sname achieves consistently larger improvements across all benchmarks.
This suggests that directly learning the future representations generated from noise is particularly effective for improving action control in WAMs.

\begin{table}[t]
\centering
\small
\begin{minipage}[t]{0.54\linewidth}
\vspace{0pt}
\centering
\caption{
\textbf{Component ablation study of FRED.} We ablate self-distillation and history conditioning on DROID in-domain long-horizon (LH) task, OpenArm human skill transfer (HS) tasks, and RoboCasa Kitchen. We report average task progress rate (\%) for DROID LH and average success rate (\%) otherwise. All 5B models are trained for 100K iterations on DROID (batch size 128), 50K iterations on OpenArm with human videos, and 100K iterations on RoboCasa Kitchen (batch size 64 for both).
}
\label{tab:method_ablation}
\begin{adjustbox}{max width=0.98\linewidth}
\begin{tabular}{l ccc}
\toprule
Method & DROID (LH) & OpenArm (HS) & RoboCasa \\
\midrule
\textbf{\sname (Ours)} & \textbf{43.3} & \textbf{50.0} & \textbf{68.8} \\
\phantom{0}w/o Self-distillation & 21.7 & 33.3 & 65.1 \\
\phantom{0}w/o History & 28.3 & 32.1 & 64.8 \\
\midrule
Base WAM (DiT4DiT) & \phantom{0}8.3 & 10.7 & 61.4 \\
\bottomrule
\end{tabular}
\end{adjustbox}
\end{minipage}\hfill%
\begin{minipage}[t]{0.44\linewidth}
\vspace{0pt}
\centering
\caption{
\textbf{Analysis of \sname design choices.}
We examine the effects of (a) teacher noise level,
(b) feature extraction layer (index), and
(c) self-distillation loss coefficient
on RoboCasa Kitchen. All 5B models are trained for 100K iterations with batch size 64.
Default settings are highlighted in \colorbox{baselinecolor}{gray}.}
\label{tab:analysis_design_choices}
\begin{subtable}[t]{0.32\linewidth}
    \centering\small
    \caption{Teacher noise}
    \label{tab:analysis_design_choices:teacher_noise}
    \vspace{0.092em}
\begin{adjustbox}{max width=0.9\linewidth}
    \begin{tabular}{cc}
    \toprule
    $\tau$ & SR (\%) \\
    \midrule
    0.00 & 65.3 \\
    \baseline{0.25} & \baseline{\textbf{68.8}} \\
    0.50 & 67.8 \\
    0.75 & 67.6 \\
    \bottomrule
    \end{tabular}
\end{adjustbox}
\end{subtable}\hfill%
\begin{subtable}[t]{0.32\linewidth}
    
    \centering
    \caption{Feature layer}\label{tab:analysis_design_choices:distill_layer}
    \vspace{0.36em}
\begin{adjustbox}{max width=0.98\linewidth}
    \begin{tabular}{cc}
    \toprule
    Layer & SR (\%) \\
    \midrule
    10 & 57.5 \\
    14 & 63.0 \\
    \baseline{18} & \baseline{\textbf{68.8}} \\
    22 & 66.8 \\
    26 & 64.7 \\
    \bottomrule
    \end{tabular}
\end{adjustbox}
\end{subtable}\hfill%
\begin{subtable}[t]{0.32\linewidth}
    
    \centering
    \caption{Loss coefficient}
    \label{tab:analysis_design_choices:sd_coeff}
    \vspace{1.36em}
\begin{adjustbox}{max width=0.92\linewidth}
    \begin{tabular}{cc}
    \toprule
    $\lambda_{\mathrm{SD}}$ & SR (\%) \\
    \midrule
    0.1   & 68.1 \\
    \baseline{0.01}  & \baseline{\textbf{68.8}} \\
    0.001 & 62.7 \\
    \bottomrule
    \end{tabular}
\end{adjustbox}
\end{subtable}
\end{minipage}
\end{table}

\begin{table}[t]
\centering
\caption{
\textbf{Comparison of representation learning methods on simulation benchmarks.}
For each benchmark, all 5B models are trained on its training dataset for 100K iterations with a batch size of 64.
We report the average success rate (\%) on each benchmark.
}
\label{tab:sim_results}

\scalebox{0.85}{
\begin{tabular}{lccc}
\toprule
Method & RoboCasa & GR-1 Tabletop & LIBERO \\
\midrule
DiT4DiT \citep{ma2026dit4dit}
& 61.4 & 50.4 & 97.4 \\
\quad w/ VideoREPA \citep{zhang2026videorepa}
& 64.3 & 51.6 & 97.3 \\
\quad w/ SRA \citep{jiang2025no}
& 62.9 & 52.1 & 97.6 \\
\quad w/ Self-Flow \citep{chefer2026self}
& 61.8 & 50.9 & 97.9 \\
\quad \textbf{w/ \sname{} (Ours)}
& \textbf{68.8} & \textbf{54.8} & \textbf{98.2} \\
\bottomrule
\end{tabular}
}
\end{table}

\begin{figure*}[t]
\centering\small
\begin{subfigure}[t]{0.31\textwidth}
  \centering
  \includegraphics[width=0.92\linewidth]{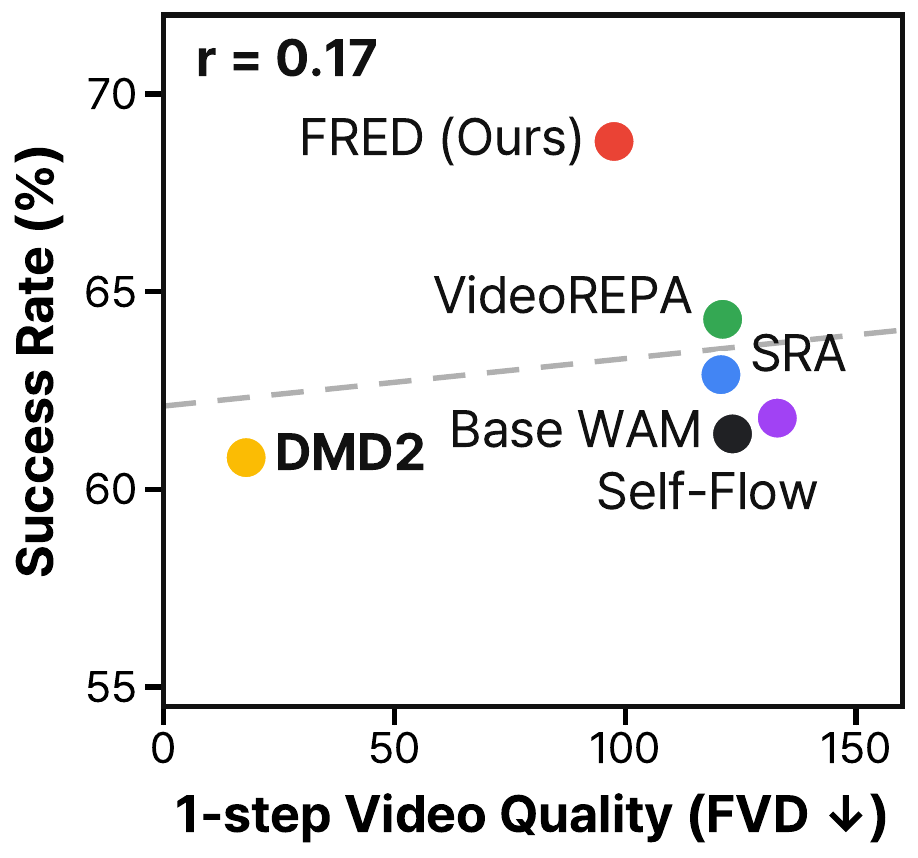}
  \subcaption{One-step video generation quality}
  \label{fig:observation:sr_vidgen}
\end{subfigure}~
\begin{subfigure}[t]{0.31\textwidth}
  \centering
  \includegraphics[width=0.92\linewidth]{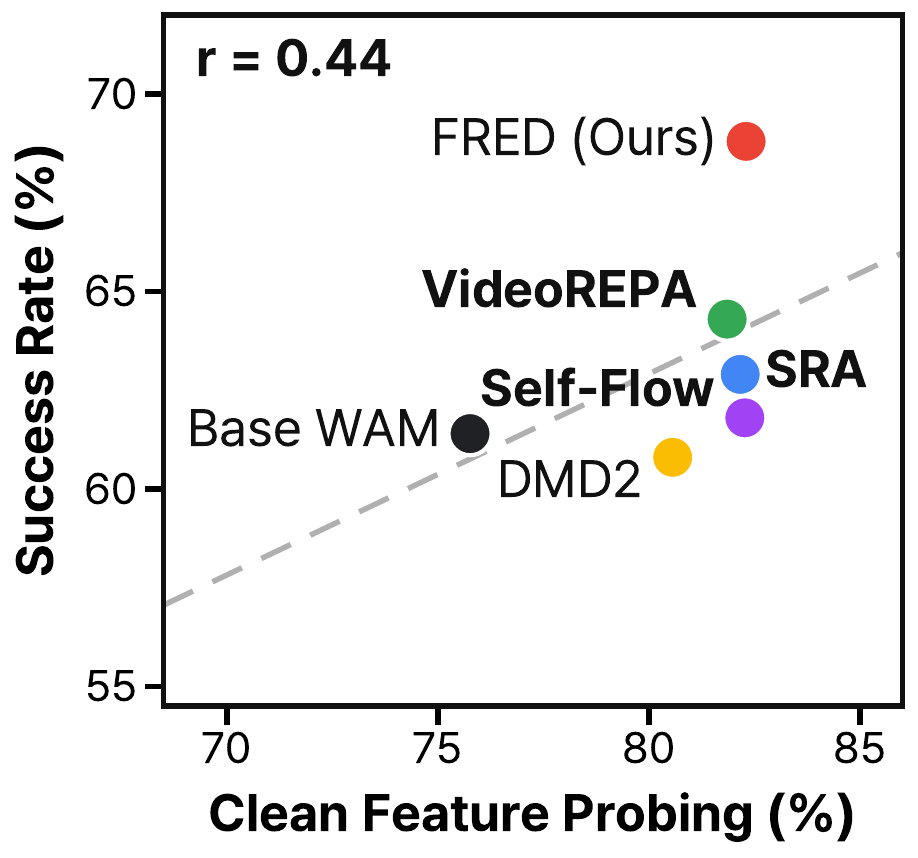}
  \subcaption{Representation quality from future observations}
  \label{fig:observation:sr_repclean}
\end{subfigure}~
\begin{subfigure}[t]{0.31\textwidth}
  \centering
  \includegraphics[width=0.92\linewidth]{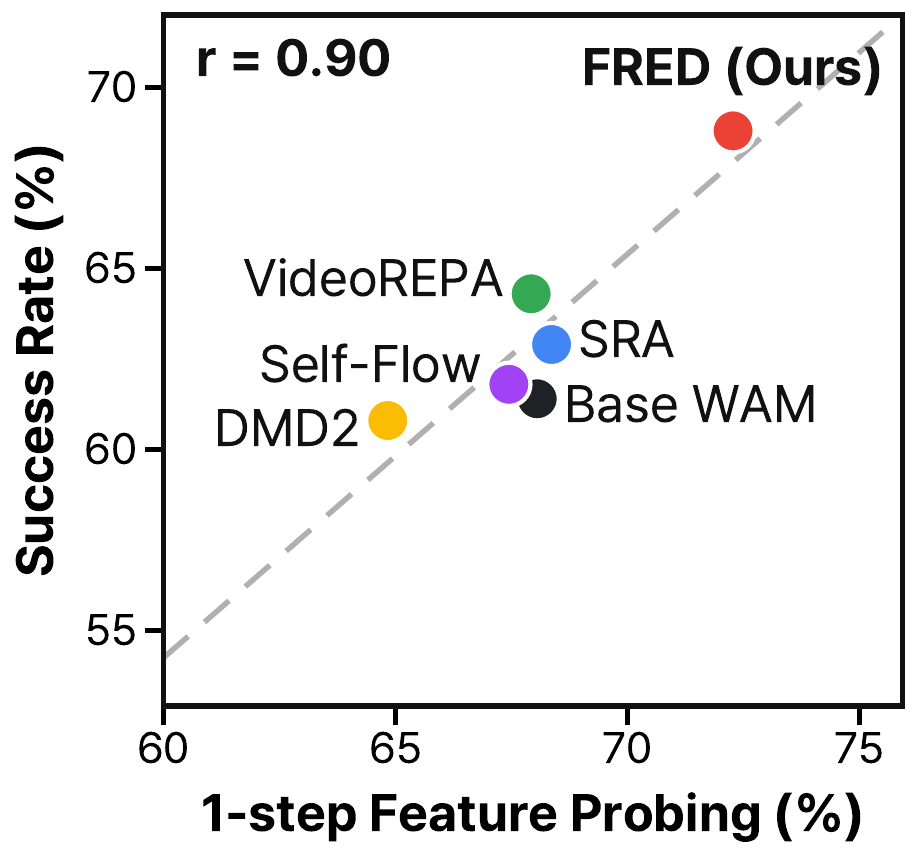}
  \subcaption{Representation quality from one-step video denoising}
  \label{fig:observation:sr_repgen}
\end{subfigure}
\caption{
\textbf{Policy performance versus video generation and representation quality.}
With DiT4DiT as the base WAM, we compare variants trained with different VDM objectives in terms of task success, video generation quality, and representation quality on RoboCasa Kitchen.
We find that the task success rates correlate weakly with (a) one-step video generation quality and (b) representation quality from clean future observations, but more strongly correlate with (c) the quality of one-step generated future representations from noise.
}
\label{fig:observations}
\end{figure*}

\textbf{Future Representation Quality and Robot Control.}
Since the action decoder directly conditions on future video representations generated from noise, their quality may be closely correlated with robot control performance.
To investigate this, we use DiT4DiT~\citep{ma2026dit4dit} as the base WAM and compare variants with different VDM training objectives along three dimensions: one-step video generation quality, representation quality on observed future videos, and the quality of future representations generated from noise.
Specifically, the WAM variants include an objective that explicitly improves one-step video generation through step-distillation (DMD2) \citep{yin2024improved} and representation learning objectives that improve intermediate video representations.
We measure video generation quality using Fr\'echet Video Distance (FVD) \cite{unterthiner2018towards}.
For the representation quality, we follow prior works on evaluating visual representations with frozen-backbone probing~\citep{yu2024representation,assran2025v}.
In particular, we train a lightweight attentive probe on the video representations to classify each video into predefined motion primitive classes (see Appendix~\ref{appendix:analysis_setup} for more detailed setup).

In \Cref{fig:observation:sr_vidgen,fig:observation:sr_repclean}, we find that task success is only weakly correlated with one-step video generation quality ($r=0.17$) and representation quality on observed future videos ($r=0.44$).
In contrast, in \Cref{fig:observation:sr_repgen}, we observe that the quality of future representations generated from noise shows a stronger correlation with task success ($r=0.90$).
This supports our view of the VDM as a future representation generator and suggests that improving the representations directly used for action generation may be particularly relevant to control performance.

\begin{table*}[t!]
\centering
\small

\begin{minipage}[t]{0.49\textwidth}
    \centering
    \captionof{table}{
    \textbf{Generalization across video backbones.}
    We evaluate \sname with different VDM backbones on RoboCasa Kitchen. All models are trained for 100K iterations with a batch size of 64.
    }
    \label{tab:backbone_generalization}
    \vspace{1.17em}
    \begin{adjustbox}{max width=\linewidth}
    \begin{tabular}{lc}
    \toprule
    Method & SR (\%) \\
    \midrule
    DiT4DiT (Wan2.2-5B) & 61.4 \\
    \textbf{\quad w/ \sname{} (Ours)} & \textbf{68.8} \\
    \midrule
    DiT4DiT (Cosmos-Predict2.5-2B) & 56.7 \\
    \textbf{\quad w/ \sname{} (Ours)} & \textbf{62.1} \\
    \bottomrule
    \end{tabular}
    \end{adjustbox}
\end{minipage}
\hfill
\begin{minipage}[t]{0.49\textwidth}
    \centering
    \captionof{table}{
    \textbf{Inference cost.}
    Average inference latency in the real-world Franka Research 3 setup on an NVIDIA A100 80GB GPU.
    }
    \label{tab:efficiency_analysis}
    \vspace{-0.5em}
    \begin{adjustbox}{max width=\linewidth}
    \begin{tabular}{lccc}
    \toprule
    Method
    & Model Size & \# Frames
    & Time (s, $\downarrow$) \\
    \midrule
    $\pi_{0.5}$
    & 3B & 1 & 0.261 \\
    GR00T N1.7
    & 3B & 1 & 0.121 \\
    \midrule
    DiT4DiT & 5B & 1 & 0.111 \\
    \textbf{\sname{} (Ours)} 
    & 5B & 5 & 0.136 \\
    \midrule
    DreamZero
    & 14B & 5 & 3.255 \\
    \textbf{\sname{} (Ours)}
    & 14B & 5 & 0.406 \\
    \bottomrule
    \end{tabular}
    \end{adjustbox}
\end{minipage}

\end{table*}

\textbf{Generality across Video Backbones.}
In \Cref{tab:backbone_generalization}, we evaluate whether the gains of \sname generalize across different video diffusion backbones.
We find that \sname consistently improves the success rates of DiT4DiT~\citep{ma2026dit4dit} with both the Wan2.2-5B~\citep{wan2025wan} and Cosmos-Predict2.5-2B~\citep{ali2025world} backbones, demonstrating that its effectiveness extends beyond a single video backbone.

\textbf{Inference Cost.}
In \Cref{tab:efficiency_analysis}, we measure the average inference latency of \sname and other baselines at different model scales. At the 5B scale, we observe that \sname achieves inference latency comparable to the VLA baselines, despite conditioning on 5 frames by default, with only a 1.2$\times$ increase over DiT4DiT with 1-frame input. At the 14B scale, \sname is approximately 8$\times$ faster than DreamZero~\citep{ye2026world}, demonstrating the practical efficiency of using a single video denoising step for action generation.

\section{Conclusion}
In this paper, we have presented \sname, a novel framework for training World Action Models (WAMs) that explicitly learns the future representations used for action generation from a single video denoising step via self-distillation.
Our key insight is to view the video diffusion model (VDM) as a future representation generator
and explicitly improve these representations by aligning them with self-distillation targets extracted from noisy future observations, while using
observation history as additional temporal context.
Our experiments show that \sname consistently improves robot control in diverse real-world scenarios, including zero-shot environment generalization and human-to-robot skill transfer.
We hope our work on the connection between representation learning and WAMs encourages further work toward efficient and generalizable robot foundation models that better leverage large-scale pre-trained video generation models.
We provide more discussion in Appendix~\ref{appendix:limitations}.

\section*{Acknowledgement}
For SJ, HJ, and JS: this paper was supported by RLWRLD Inc. This research was supported by the ``Advanced GPU Utilization Support Program'' funded by the Government of the Republic of Korea (Ministry of Science and ICT).
We thank Myungkyu Koo, Beomjun Kim, Junhyeong Park, Junyoung Sung, Seunghoon Sim, John Won, Seungyong Lee, and Wook Jung for their help in collecting robot and human demonstrations for our real-world experiments. SY thanks Shusheng Yang for providing the manuscript template.

\addcontentsline{toc}{section}{References}
\bibliography{references}
\bibliographystyle{references}

\clearpage
\newpage
\appendix

\crefalias{section}{appendix}
\crefalias{subsection}{subappendix}
\crefalias{subsubsection}{subsubappendix}

\section{Implementation Details}\label{appendix:implementation_details}

\textbf{Implementation Details for Real-world Experiments.}
For the Franka Research 3 experiments, we consider two different model scales, where we use Wan2.2-TI2V-5B and Wan2.1-I2V-14B as the 5B and 14B video diffusion backbones, respectively, while all OpenArm experiments use Wan2.2-TI2V-5B.
We condition the video model on 4 historical frames with the current frame, and predict 8 future frames, with a temporal stride of 3.
For \sname, we use the same intermediate layer for self-distillation and action conditioning (19th layer for Wan2.2-5B and 25th layer for Wan2.1-14B). Across all experiments, we use $\gamma_1=0.1$, $\gamma_2=0.04$, teacher noise level $\tau=0.25$, and $\lambda_{\mathrm{SD}}=0.01$.
We update the teacher with an exponential moving average (EMA) using the $1-(1+t)^{-3/4}$ warmup decay schedule implemented in Diffusers.
We normalize actions and proprioceptive states using Q01-Q99 normalization.

We train the Franka 5B and 14B models for 150K and 100K iterations with batch size 128.
OpenArm models are trained for 50K iterations; robot-only training uses batch size 32, while robot-human co-training uses 32 robot and 32 human samples per batch.
All models are optimized with AdamW and cosine learning rate decay. For the 14B \sname model trained on DROID, we set the video model learning rate to $10^{-5}$, following DreamZero. For all other settings, we use a learning rate of $10^{-4}$ for both the video model and action decoder.
During training, we sample video diffusion timesteps from a shifted-uniform distribution, following the official Wan2.1 and Wan2.2 implementations~\citep{wan2025wan}, and sample action diffusion timesteps from a $\mathrm{Beta}(1.5, 1.0)$ distribution. At inference, we generate actions using 4 denoising steps.

For baseline comparisons, we use publicly released checkpoints if available. Otherwise, we train them following their official implementations, matching the number of training iterations and batch size to our corresponding experiments.
Specifically, for DROID experiments, we use the publicly released checkpoints for $\pi_{0.5}$, GR00T N1.7, and DreamZero. For DreamZero on OpenArm, Fast-WAM, and DiT4DiT, we follow the official training recipes. AGRA is reproduced in our codebase since there is no available official implementation.
We provide the detailed training configurations of \sname in \Cref{tab:training_details}.

\begin{table*}[h]
\centering
\caption{\textbf{Hyperparameter details for training \sname.} FR3 denotes Franka Research 3.}
\label{tab:training_details}

\begin{adjustbox}{max width=0.95\textwidth}
\begin{tabular}{l cc c cc}
\toprule
& \multicolumn{2}{c}{FR3 (\Cref{tab:droid_real})}
& FR3 (\Cref{tab:method_ablation})
& \multicolumn{2}{c}{OpenArm (\Cref{tab:openarm_real,tab:method_ablation})} \\
\cmidrule(lr){2-3}\cmidrule(lr){4-4}\cmidrule(lr){5-6}
& Wan2.2-5B & Wan2.1-14B & Wan2.2-5B & w/o human data & w/ human data \\
\midrule
\multicolumn{6}{l}{\textbf{Video Architecture}} \\
Video backbone & Wan2.2-TI2V-5B & Wan2.1-I2V-14B & Wan2.2-TI2V-5B & Wan2.2-TI2V-5B & Wan2.2-TI2V-5B \\
History / future frames & 5 / 8 (stride 3) & 5 / 8 (stride 3) & 5 / 8 (stride 3) & 5 / 8 (stride 3) & 5 / 8 (stride 3) \\
Input resolution & $288\times320$ & $288\times320$ & $288\times320$ & $384\times256$ & $384\times256$ \\
\midrule
\multicolumn{6}{l}{\textbf{Action Architecture}} \\
Num. layers & 16 & 24 & 16 & 16 & 16 \\
Hidden dim. & 768 & 1024 & 768 & 768 & 768 \\
Num. heads & 12 & 16 & 12 & 12 & 12 \\
Action horizon & 24 & 24 & 24 & 24 & 24 \\
\midrule
\multicolumn{5}{l}{\textbf{Optimization}} \\
Training iterations & 150K & 100K & 100K & 50K & 50K \\
Warmup iterations & 5K & 5K & 5K & 2.5K & 2.5K \\
Global batch size & 128 & 128 & 128 & 32 & 64 (robot 32 + human 32) \\
Optimizer & AdamW & AdamW & AdamW & AdamW & AdamW \\
Learning rate (video / action) & $10^{-4}$ / $10^{-4}$ & $10^{-5}$ / $10^{-4}$ & $10^{-4}$ / $10^{-4}$ & $10^{-4}$ / $10^{-4}$ & $10^{-4}$ / $10^{-4}$ \\
$(\beta_1,\beta_2)$ & $(0.9, 0.95)$ & $(0.95, 0.999)$ & $(0.9, 0.95)$ & $(0.9, 0.95)$ & $(0.9, 0.95)$ \\
Weight decay & $10^{-2}$ & $10^{-5}$ & $10^{-2}$ & $10^{-2}$ & $10^{-2}$ \\
Learning rate schedule & Cosine decay & Cosine decay & Cosine decay & Cosine decay & Cosine decay \\
Minimum learning rate ratio & 0.1& 0.1 & 0.1 & 0.1 & 0.1 \\
\bottomrule
\end{tabular}
\end{adjustbox}
\end{table*}

\clearpage

\textbf{Implementation Details for Simulation Experiments.}
For each simulation benchmark in \Cref{tab:sim_results}, we train \sname with the same Wan2.2-TI2V-5B~\citep{wan2025wan} backbone and 16-layer DiT action decoder for 100K iterations with a global batch size of 64. 
We use three concatenated views (two external and one wrist) at $384\times256$ for RoboCasa Kitchen~\citep{nasiriany2024robocasa}, one egocentric view at $256\times256$ for GR-1 Tabletop~\citep{bjorck2025gr00t}, and two concatenated views (one external and one wrist) at $256\times512$ for LIBERO~\citep{liu2023libero}.
We fix the action prediction horizon at 16 across all benchmarks, and we normalize proprioceptive states and actions using min-max normalization for RoboCasa Kitchen and LIBERO, and Q01-Q99 normalization for GR-1 Tabletop. We condition the video model on 4 historical frames and the current frame, and predict 8 future frames, with a temporal stride of 2.

All models are optimized with AdamW using cosine learning rate decay and a 5K iteration warmup. We set the peak learning rate to $10^{-4}$ for both the video model and action decoder, while the minimum learning rate is 1\% of the peak across all benchmarks. For the remaining \sname-specific hyperparameters and training and inference details, we follow those used in the real-world experiments.

For baseline comparisons, we follow the official implementation of each method if it is available. We use DiT4DiT~\citep{ma2026dit4dit} as the base WAM and make one change to its official implementation for simulation experiments. We condition the action decoder only on future video latent tokens, whereas the original implementation uses both current and future video latent tokens (see \Cref{tab:appendix_action_condition_token} for an ablation of this design choice). We build \sname and all representation learning baselines on this modified DiT4DiT implementation.
Specifically, for VideoREPA~\citep{zhang2026videorepa}, we follow its official implementation with a VideoMAEv2 ViT-B encoder~\citep{wang2023videomae} to produce the alignment target, modifying only the encoder input to match our input resolution and prediction horizon. For SRA~\citep{jiang2025no}, we follow its official SiT-XL/2~\citep{ma2024sit} implementation for video representation alignment. For Self-Flow~\citep{chefer2026self}, since there is no available official implementation for training, we reproduce the alignment procedure and layer selection described in the original paper.

\textbf{Augmentations.} We summarize the visual augmentations used in our experiments in \Cref{tab:image_augmentation}.

\begin{table*}[h]
\centering
\caption{\textbf{PyTorch-like descriptions of image augmentations for \sname.}}
\label{tab:image_augmentation}
\begin{adjustbox}{max width=0.8\textwidth}
\begin{tabular}{cl}
\toprule
Experiment & Descriptions \\
\midrule
\multirow{6}{*}{LIBERO} 
& \texttt{Resize((H, W))} \\
& \texttt{RandomResizedCrop((H, W), scale=(0.9, 0.9), ratio=(W/H, W/H))} \\
& \texttt{RandomBrightness(scale=(0.8, 1.2))} \\
& \texttt{RandomContrast(scale=(0.8, 1.2))} \\
& \texttt{RandomSaturation(scale=(0.8, 1.2))} \\
& \texttt{RandomHue(scale=(-0.05, 0.05))} \\
\midrule
\multirow{7}{*}{All others}
& \texttt{Resize((H, W))} \\
& \texttt{RandomResizedCrop((H, W), scale=(0.9, 0.9), ratio=(W/H, W/H))} \\
& \texttt{RandomRotate(scale=(-5$^\circ$, 5$^\circ$))} \\
& \texttt{RandomBrightness(scale=(0.8, 1.2))} \\
& \texttt{RandomContrast(scale=(0.8, 1.2))} \\
& \texttt{RandomSaturation(scale=(0.8, 1.2))} \\
& \texttt{RandomHue(scale=(-0.05, 0.05))} \\
\bottomrule
\end{tabular}
\end{adjustbox}
\end{table*}

\newpage

\section{Benchmark Details}\label{appendix:benchmark_details}
\subsection{Zero-shot Environment Generalization Tasks on DROID}
\label{appendix:benchmark_details:droid}

\begin{figure}[h]
    \centering
    \begin{subfigure}[t]{0.48\linewidth}
        \centering
        \includegraphics[width=\linewidth]{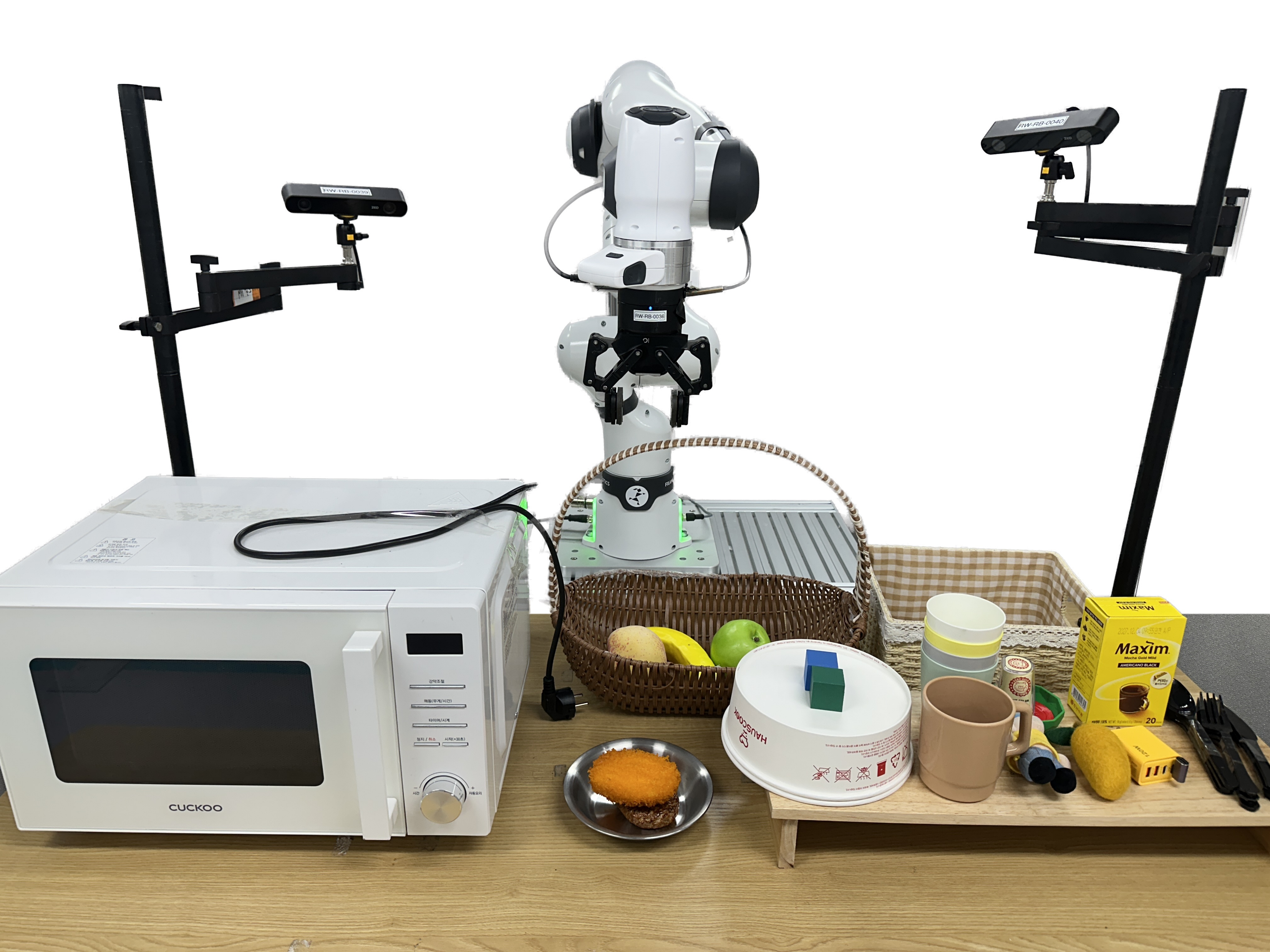}
        \subcaption{Env.~1 (wooden tabletop)}
        \label{fig:droid_setup:env1}
    \end{subfigure}\hfill
    \begin{subfigure}[t]{0.48\linewidth}
        \centering
        \includegraphics[width=\linewidth]{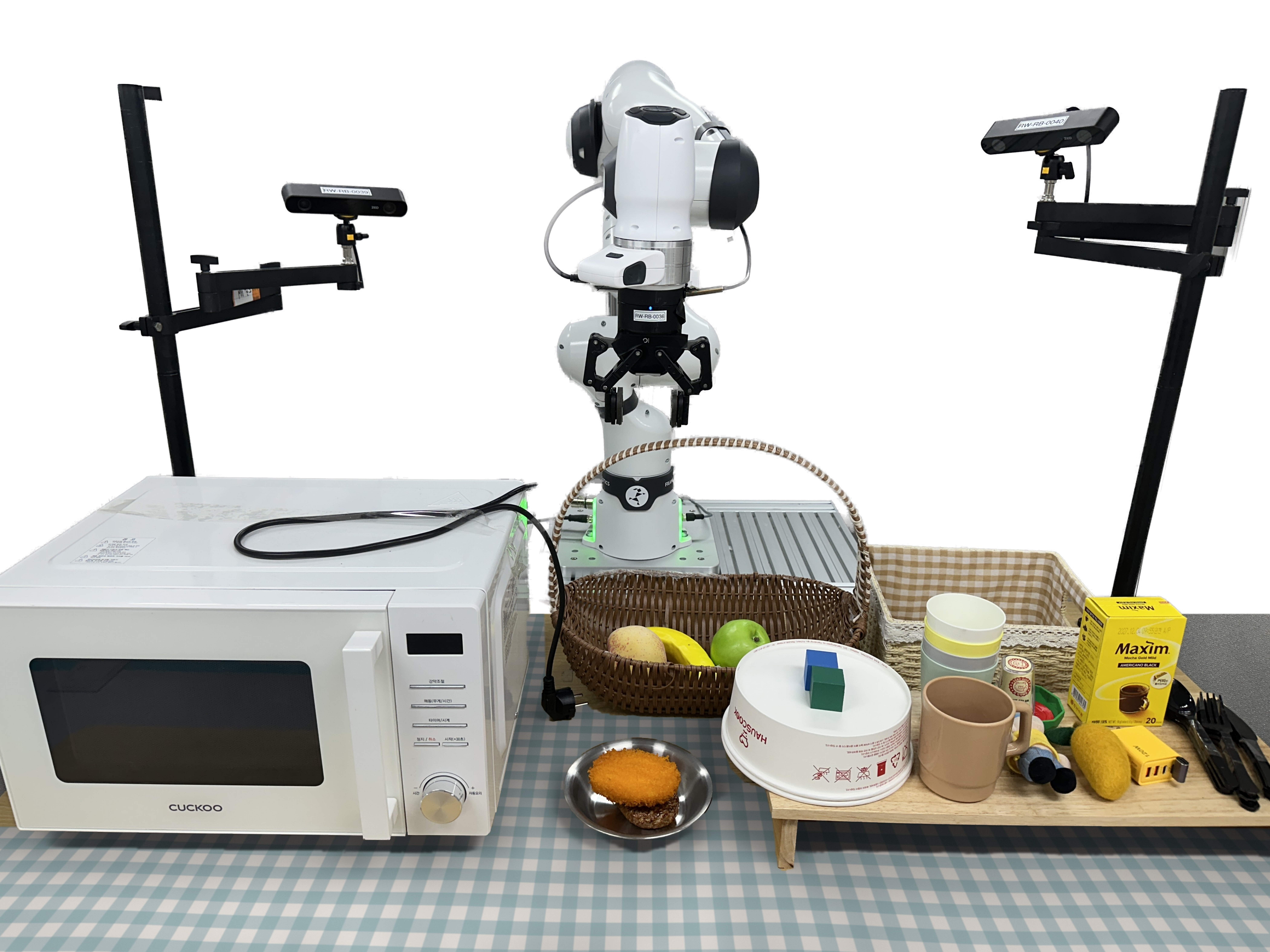}
        \subcaption{Env.~2 (blue checkered tablecloth)}
        \label{fig:droid_setup:env2}
    \end{subfigure}
    
    \caption{\textbf{DROID evaluation setup.} The setup includes a Franka Research 3 robot arm equipped with a Robotiq 2F-85 gripper, a wrist-mounted ZED Mini camera, two external ZED 2 cameras (left and right), and a variety of task objects. The two environments feature (a) an exposed wooden tabletop and (b) a tabletop covered with a blue checkered tablecloth.}
    \label{fig:droid_setup}
\end{figure}

\textbf{Hardware Setup.}
We evaluate models trained on DROID across two new environments without additional training on data from either environment. These environments differ from those in the training data in location, background, and objects. One has an exposed wooden tabletop, and the other has a blue checkered tablecloth (see Env. 1 and Env. 2 in \Cref{tab:droid_real}, respectively). We conduct these experiments using a Franka Research 3 robot arm equipped with a Robotiq 2F-85 gripper, along with a wrist-mounted ZED Mini stereo camera and two external ZED 2 stereo cameras. The action space consists of relative joint angles (7D) and an absolute gripper opening (1D). For $\pi_{0.5}$~\citep{intelligence2025pi_} and GR00T N1.7~\citep{nvidia2026gr00tn17}, we follow the official implementations that use the wrist camera and the left external camera. For the remaining models, we use the wrist camera and both the left and right external cameras. We show the hardware setup in \Cref{fig:droid_setup}.

\newpage

\begin{figure}[h]
    \centering
    \includegraphics[width=0.98\textwidth]{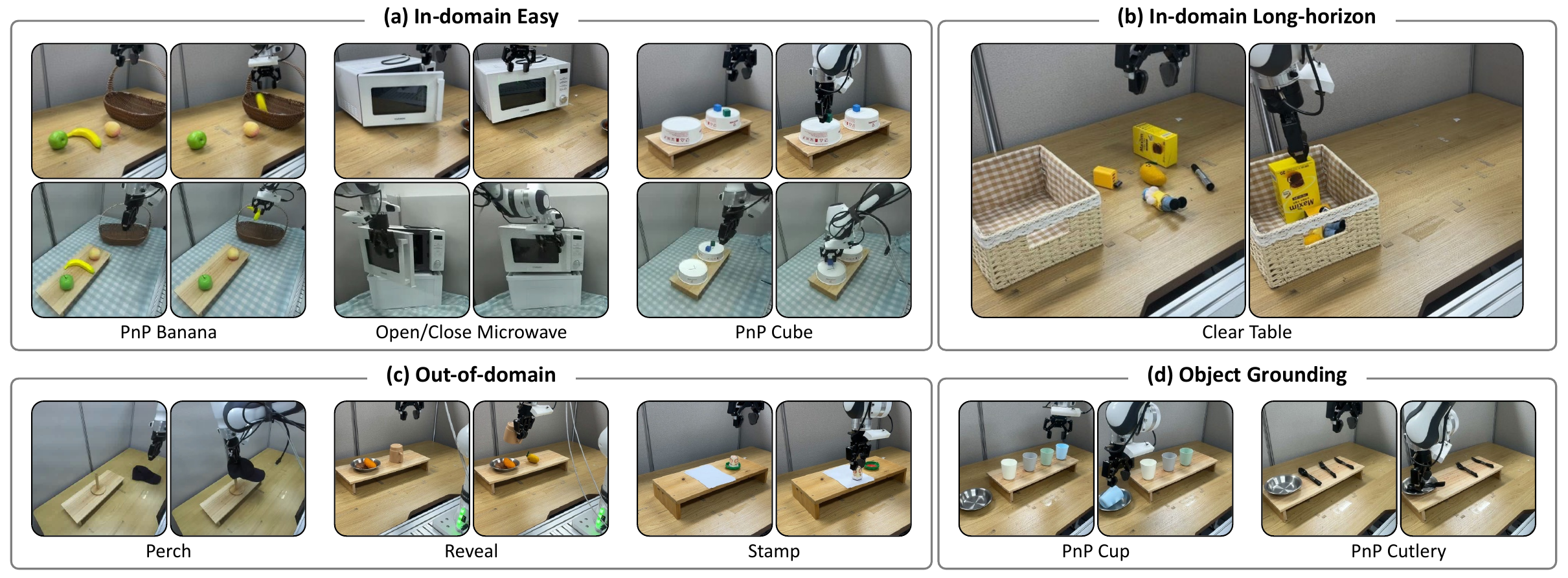}
    \vspace{-0.5em}
    \caption{\textbf{DROID evaluation tasks.} Task visualization for 4 task suites: (a) in-domain easy, (b) in-domain long-horizon, (c) out-of-domain, and (d) object grounding. \textit{Clear Table} repeats the pick-and-place step until the table is empty.}
    \label{fig:all_tasks}
\end{figure}

\textbf{Evaluation Details.}
We use the environment with the exposed wooden tabletop as our default evaluation setting (Env. 1 in \Cref{fig:droid_setup}). In this setting, we evaluate four task sets covering (a) \textit{in-domain easy tasks}, (b) \textit{in-domain long-horizon task}, (c) \textit{out-of-domain tasks}, and (d) \textit{object grounding tasks}. We consider a task in-domain if its action verb is present in the DROID dataset, and out-of-domain otherwise. The object grounding tasks test whether the policy can distinguish objects with similar colors or shapes. In the default environment, we evaluate each task over 12 trials and report task progress rates (\%) for the in-domain long-horizon task and binary success rates (\%) for all other tasks. For each in-domain easy task, we run 12 additional trials in the environment with the blue checkered tablecloth (Env. 2 in \Cref{fig:droid_setup}).  We limit each trial to 1,500 action steps for the long-horizon task and 500 action steps for all other tasks. We describe the task configurations below (see \Cref{fig:all_tasks} for visualization of the tasks).
\begin{itemize}[leftmargin=*,itemsep=0mm]
\vspace{-0.5em}
\item \textbf{(In-domain Easy) PnP Banana.} The policy must pick up the banana and place it in the basket. We place the basket on either the left or the right. We place an apple and a peach near the banana as distractors. We use the instruction \textit{``Put the banana in the basket''} for the 14B FRED model and WAM baselines and \textit{``Pick up the banana and place it in the basket''} for the 5B FRED model, WAM baselines and VLAs.
\item \textbf{(In-domain Easy) Open/Close Microwave.} The policy must open or close the microwave door as instructed. We vary both the action and microwave position (left/center/right).
\item \textbf{(In-domain Easy) PnP Cube.} The policy must move the specified cube to the specified plate. We vary the cube color between green and blue and the target plate between left and right. The instruction is \textit{``Move the (green/blue) cube to the (left/right) plate''.}
\item \textbf{(In-domain Long-horizon) Clear Table.} The policy must pick up all 5 objects on the table and place them in the box. We use the instruction \textit{``Pick up all the objects on the table and put them in the box''.} We give 0.2 points for each object successfully placed in the box.
\item \textbf{(Out-of-domain) Perch.} The policy must pick up the hat from the table and fit it over the upright wooden peg. We position the hat and peg diagonally across from each other on the table and mirror their positions left to right across trials. We use the instruction \textit{``Perch the hat on the upright wooden peg''.}
\item \textbf{(Out-of-domain) Reveal.} The policy must lift the mug that covers an object and expose the object underneath. We place distractor objects to the left and to the right of the mug across trials. We use the instruction \textit{``Reveal the object under the mug''.}
\item \textbf{(Out-of-domain) Stamp.} The policy must pick up the stamp and press it onto the paper to leave a mark. We use the instruction \textit{``Pick up the stamp and stamp on the paper''}.
\item \textbf{(Object Grounding) PnP Cup.} The policy must select a cup of the instructed color from four identically sized cups (white, gray, blue, and green) and place it on the plate. We vary the target color using the instruction \textit{``Pick up the (white/gray/blue/green) cup and place it on the plate''.}
\item \textbf{(Object Grounding) PnP Cutlery.} The policy must distinguish fine-grained visual differences among plastic spoons, knives, and forks, and place the instructed one
on the plate. We vary the target piece using the instruction \textit{``Pick up the (spoon/knife/fork) and place it on the plate''.}
\end{itemize}

\newpage
\subsection{Human-robot Co-training on OpenArm}
\label{appendix:benchmark_details:openarm}

\begin{figure*}[h]
    \centering
    \includegraphics[width=0.7\textwidth]{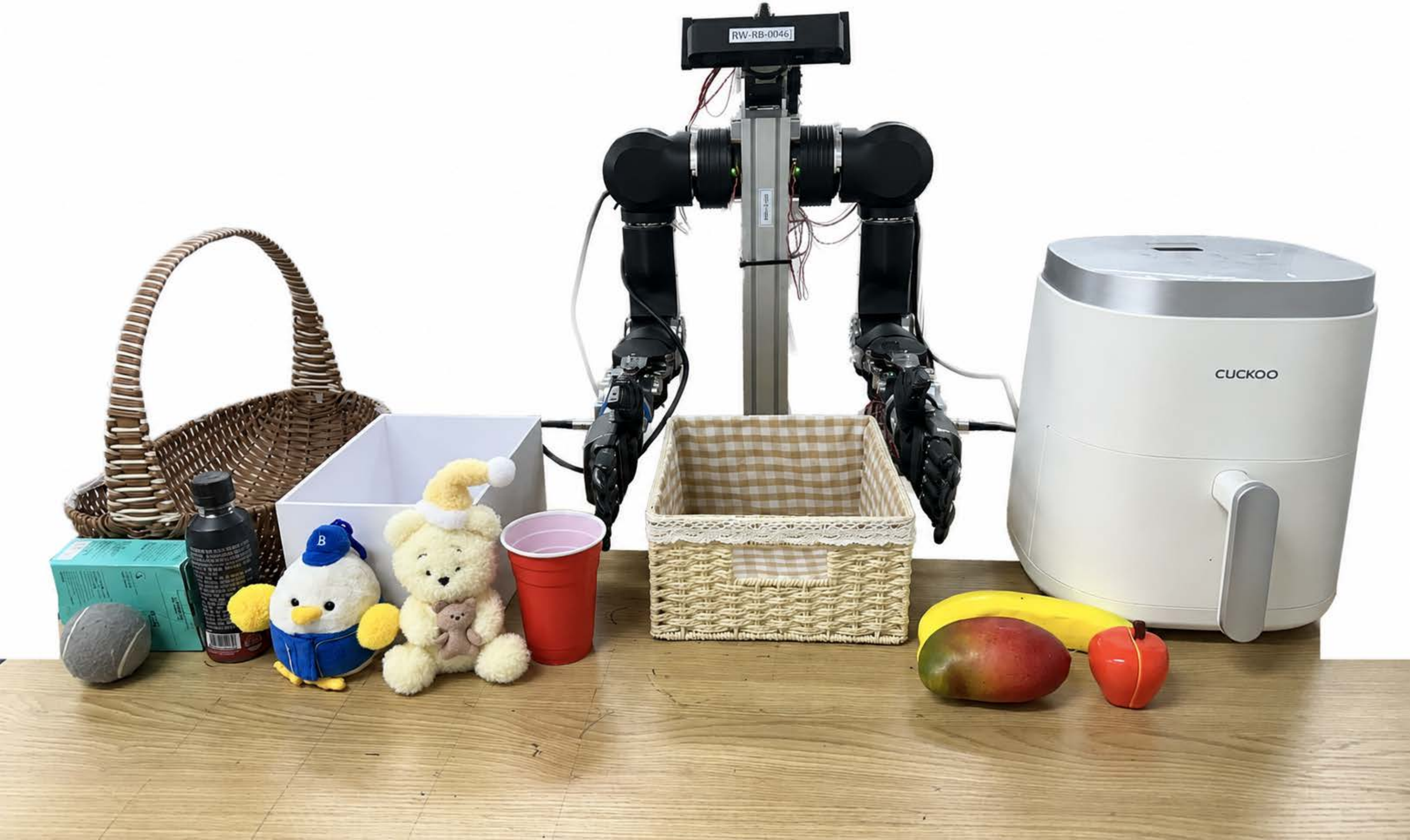}
    \caption{\textbf{OpenArm evaluation setup.} It consists of the two 7-DoF OpenArm arms and a ZED 2i stereo camera mounted on the 2-DoF neck, together with the objects used in the evaluation tasks.}
    \label{fig:openarm_setup}
\end{figure*}

\textbf{Hardware Setup.}
We use the OpenArm humanoid\footnote{https://openarm.dev/}, which consists of two 7-DoF arms. Each arm is equipped with a 6-DoF Inspire RH56F1 hand, while the head is mounted on a 2-DoF neck supporting pitch and yaw motions.
We use relative joint actions for the arms, absolute actions for the hands, and absolute actions for the neck.
A ZED 2i stereo camera is mounted on the neck, and we use its left and right egocentric views as visual inputs in all experiments.
For data collection, we teleoperate the two arms using ExoArm-7, a wearable exoskeleton compatible with OpenArm. Finger motions are captured using Manus Pro gloves, which provide fingertip positions that are converted to finger joint angles through inverse kinematics. We visualize the hardware setup in \Cref{fig:openarm_setup}.

\newpage

\begin{figure}[h]
    \centering    \includegraphics[width=0.9\linewidth]{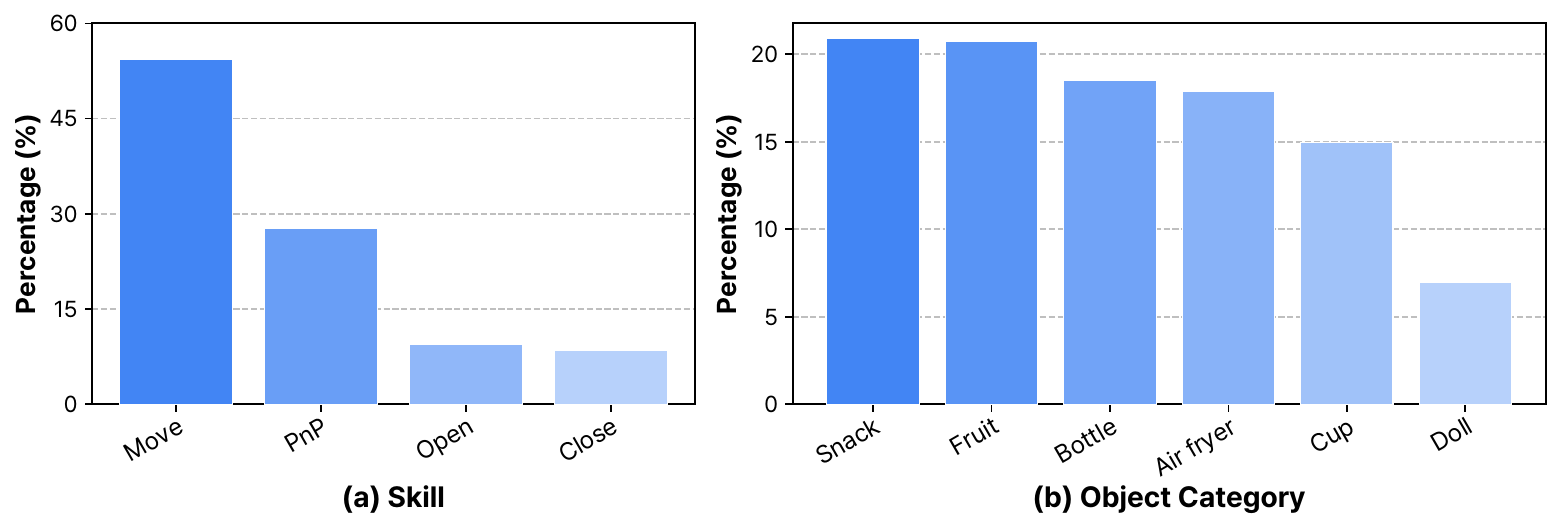}
    \vspace{-0.5em}
    \caption{\textbf{Distribution of skills and object categories in the OpenArm robot dataset.}} \label{fig:appendix:openarm_dist_robot}
\end{figure}

\begin{figure}[h]
    \centering    \includegraphics[width=0.9\linewidth]{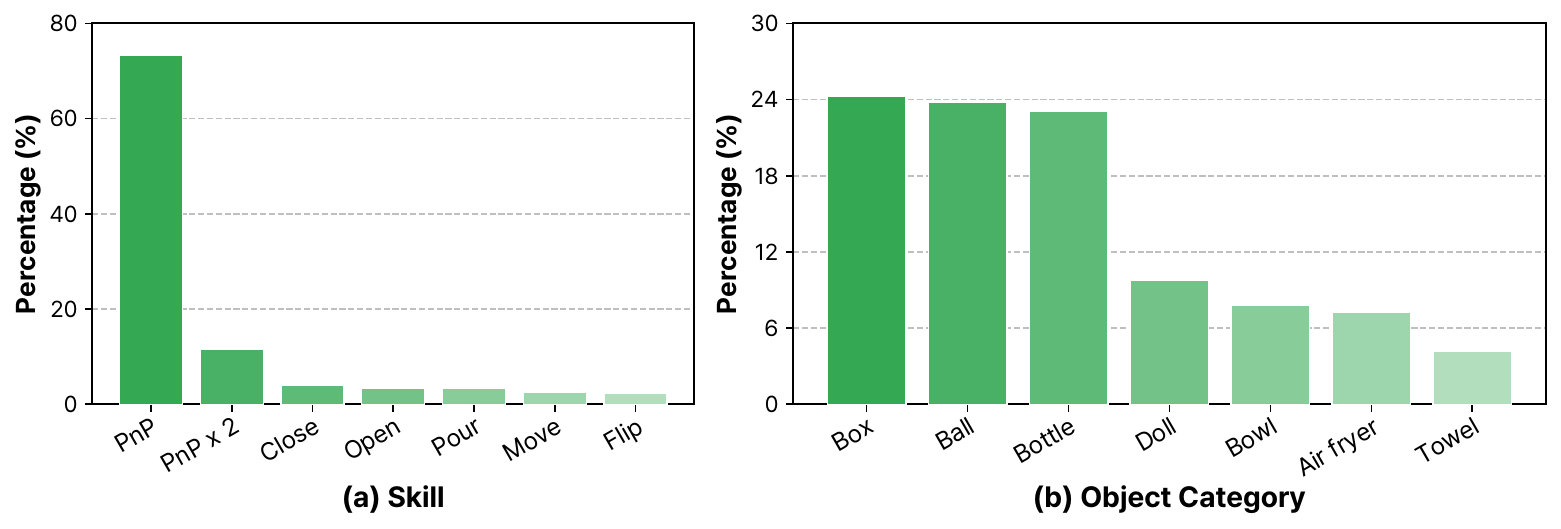}
    \vspace{-0.5em}
    \caption{\textbf{Distribution of skills and object categories in the human video dataset.}} \label{fig:appendix:openarm_dist_human}
\end{figure}

\textbf{Training Dataset.}
For the robot dataset, we collect 553 teleoperated demonstrations on the OpenArm platform, comprising 192K frames. The dataset covers 53 language instructions across 4 skills and 6 object categories; see \Cref{fig:appendix:openarm_dist_robot} for the distributions over skills and object categories.
For human co-training, we additionally collect 1,187 egocentric human videos comprising 321K frames, covering 7 skills and 7 object categories. Specifically, most data in the human dataset consists of PnP tasks, but it also includes skills absent from the robot demonstrations, such as pouring and flipping. Moreover, the human videos include object categories not present in the robot dataset, including boxes, balls, bowls, and towels, while sharing several categories such as bottles and air fryers; see \Cref{fig:appendix:openarm_dist_human} for the distributions over skills and object categories.

\newpage

\begin{figure*}[h]
    \centering
    \includegraphics[width=0.9\textwidth]{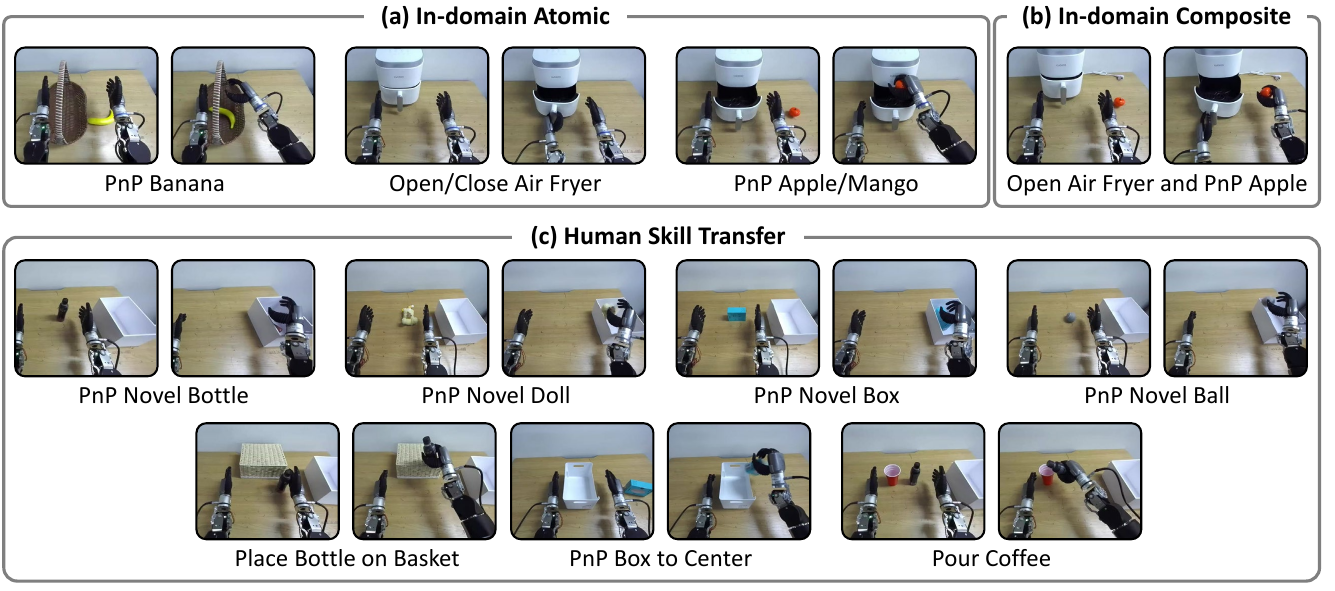}
    \vspace{-0.5em}
    \caption{\textbf{OpenArm evaluation tasks.} The tasks are grouped into (a) in-domain atomic, (b) in-domain composite, and (c) human skill transfer. We visualize initial and final egocentric frames for each task.}
    \label{fig:openarm_all_tasks}
\end{figure*}

\textbf{Evaluation Details.}
We evaluate three task sets: (a) \emph{in-domain atomic tasks}, (b) \emph{in-domain composite task}, and (c) \emph{human skill transfer tasks}.
We evaluate each task over 12 trials and report the task progress rates (\%) for the in-domain composite task, and binary success rates (\%) for other tasks. We also report instruction-following rates (\%) for human skill transfer tasks, where we give a point when the policy mimics the human-demonstrated videos, regardless of whether it successfully completes the task. We describe the task configurations below (see \Cref{fig:openarm_all_tasks} for visualization of the tasks).
\begin{itemize}[leftmargin=*,itemsep=0mm]
\vspace{-0.5em}
\item \textbf{(In-domain Atomic) PnP Banana.} The policy must place the banana in a basket. We use 6 initial positions with 2 orientations each. We use the instruction \textit{``Place the banana into the basket''.}
\item \textbf{(In-domain Atomic) Open/Close Air Fryer.} The policy must open or close an air fryer. We conduct 6 opening trials and 6 closing trials. We use the instruction \textit{``(Open/Close) the air fryer''.}
\item \textbf{(In-domain Atomic) PnP Apple/Mango.} The policy must place an apple or a red mango in the air fryer (6 trials per fruit), using the instruction \textit{``Pick up the (apple/red mango) and place it in the air fryer''.}
\item \textbf{(In-domain Composite) Open Air Fryer and PnP Apple.} The policy must open an air fryer and place an apple inside. We use the instruction \textit{``Open the air fryer, pick up the apple and place it in the air fryer''.}
\item \textbf{(Human Skill Transfer) PnP Novel Bottle.} 
The policy must place a bottle in a white container. While the bottle category is seen in robot demonstrations, the bottle instance used in this task is unseen. We use the instruction \textit{``Pick up the bottle and place it in the white container''.}
\item \textbf{(Human Skill Transfer) PnP Novel Doll.} The policy must place a doll in a white container, where the doll category is seen in robot demonstrations but the instance is unseen. We use two doll instances and conduct 6 trials per doll. We use the instruction \textit{``Pick up the doll and place it in the white container''.}
\item \textbf{(Human Skill Transfer) PnP Novel Box.} The policy must place a box, whose category is unseen in robot demonstrations, in a white container. We use the instruction \textit{``Pick up the box and place it in the white container''.}
\item \textbf{(Human Skill Transfer) PnP Novel Ball.} The policy must place a ball, whose category is unseen in robot demonstrations, in a white container. We use the instruction \textit{``Pick up the ball and place it in the white container''.}
\item \textbf{(Human Skill Transfer) Place Bottle on Basket.} The policy must place a bottle on an upside-down basket rather than inside another basket. Robot demonstrations only include placing objects inside containers, making this placement behavior unseen in robot demonstrations. We use the instruction \textit{``Pick up the bottle and put it on the basket''.}
\item \textbf{(Human Skill Transfer) PnP Box to Center.} The policy must place a box in a white container positioned between the two robot arms. Robot demonstrations do not include placing objects in containers at this location. We use the instruction \textit{``Pick up the box and place it in the white container''.}
\item \textbf{(Human Skill Transfer) Pour Coffee.} The policy must grasp and tilt the coffee bottle over the red cup and maintain the tilted position. Pouring skill is unseen in robot demonstrations. We use the instruction \textit{``Pour the coffee from the bottle into the red cup''.}
\end{itemize}

\newpage
\subsection{Simulated Robotic Manipulation Benchmarks} \label{appendix:benchmark_details:sim}
For simulated robotic manipulation experiments, we use the following benchmarks for analysis and ablation studies.

\vspace{0.5em}
\noindent\textbf{RoboCasa Kitchen} \citep{nasiriany2024robocasa} includes 24 tasks in simulated kitchen environments, including 8 pick-and-place, 6 door/drawer manipulation, and 10 appliance control tasks, following GR00T N1 \citep{bjorck2025gr00t}. We randomly sample 300 demonstrations per task in the machine-generated training dataset in RoboCasa and combine all these demonstrations to train the models. 
We evaluate models 50 times for each task with random seeds, and report the average success rates (\%). For environment initialization, the kitchen scene is selected from 5 fixed layout and style pairs, object instances are sampled from split B of RoboCasa’s object assets, and object placements and robot initial positions are randomized.

\vspace{0.5em}
\noindent\textbf{GR-1 Tabletop} \citep{bjorck2025gr00t} includes 24 tabletop tasks built upon the RoboCasa simulation framework, including 18 object rearrangement tasks and 6 articulated object manipulation tasks, following GR00T N1 \citep{bjorck2025gr00t}. We sample 1,000 demonstrations per task in the machine-generated training dataset in GR-1 Tabletop and combine all these demonstrations to train the models.
We evaluate models 50 times for each task with random seeds, and report the average success rates (\%). For environment initialization, the layout is determined by the task, while visual styles and object/container placements are randomized. For the object rearrangement tasks, object instances are sampled from split A of RoboCasa’s object assets.

\vspace{0.5em}
\noindent\textbf{LIBERO} \citep{liu2023libero} consists of 4 different sub-benchmarks (Spatial, Object, Goal, and Long), each comprising 10 tasks.
We follow the setup of $\pi_0$ \citep{black2024pi_0} that combines all training datasets in sub-benchmarks to train the models. 
We evaluate all 40 tasks across four suites (Spatial, Object, Goal, and Long), with 10 tasks per suite and 50 trials per task, and report the average success rate (\%).
For environment initialization, object positions and orientations are sampled within the task-defined regions.

\section{Baselines}\label{appendix:baselines}
We consider vision-language-action models and video diffusion-based models.
\begin{itemize}[leftmargin=*,itemsep=0.3em]
\vspace{-0.5em}
\item \textbf{$\pi_{0.5}$}~\citep{intelligence2025pi_} is a diffusion-based Vision-Language-Action model (VLA) that combines a vision-language backbone with a flow-matching action transformer. It shares self-attention layers between
the VLM and the diffusion transformer.
\item \textbf{GR00T N1.7}~\citep{nvidia2026gr00tn17} is also a diffusion-based VLA, but it feeds the intermediate hidden states of the VLM into the cross-attention layer of the diffusion transformer.
\item \textbf{DreamZero}~\citep{ye2026world} is a World Action Model (WAM) built on a pre-trained video diffusion model that jointly predicts future videos and robot actions through a single backbone.
\item \textbf{DiT4DiT}~\citep{ma2026dit4dit} is a WAM that uses separate video and action diffusion transformers, where intermediate denoising features from the video transformer condition the action transformer through cross-attention without decoding future frames.
\item \textbf{Fast-WAM}~\citep{yuan2026fast} is also a WAM that uses separate video and action diffusion transformers, while using attention masking to prevent action prediction from depending on future video tokens, thereby omitting future video prediction at inference.
\item \textbf{AGRA}~\citep{qiu2026making} extends a separate video-action architecture with an action-grounded representation alignment objective that aligns intermediate video diffusion features with spatially coherent features from a frozen visual encoder, while jointly updating the video backbone with the robot action objective.
\end{itemize}

\newpage
\section{Analysis Setup Details} \label{appendix:analysis_setup}

We describe the analysis setups used in \Cref{fig:motivation,fig:observations}.

\subsection{Attentive Probe}
\textbf{Motion-primitive Labeling.}
We partition each trajectory into non-overlapping 16-frame clips and categorize each clip into one of 15 motion-primitive labels, following \citet{zawalski2024robotic}. We determine the label from the dominant accumulated end-effector motion over the clip, with gripper state changes taking priority (see \Cref{tab:appendix_motion_primitive_distribution} for the class distribution).

\begin{table}[h]
\centering
\small
\caption{\textbf{Distribution of motion-primitive labels in the probing validation dataset.} CW and CCW denote clockwise and counterclockwise rotation, respectively.}
\label{tab:appendix_motion_primitive_distribution}
\vspace{-0.07in}
\begin{tabular}{lc lc lc}
\toprule
Primitive & (\%) & Primitive & (\%) & Primitive & (\%) \\
\midrule
Forward        & 21.82 & Down          & 13.48 & Left          & 12.91 \\
Right          & 12.27 & Backward      & 10.22 & Up            & \phantom{0}9.64 \\
Stop           & \phantom{0}5.16 & Close gripper & \phantom{0}4.17 & Open gripper  &  \phantom{0}3.11 \\
Roll right     & \phantom{0}2.39 & Tilt up       & \phantom{0}1.26 & Rotate CW     & \phantom{0}1.32 \\
Roll left      & \phantom{0}0.76 & Tilt down     & \phantom{0}0.70 & Rotate CCW    & \phantom{0}0.80 \\
\bottomrule
\end{tabular}
\end{table}

\textbf{Attentive Probe Architecture.}
Following recent self-supervised learning literature, we use a lightweight attentive probe on top of the frozen video representations \citep{bardes2024revisiting,assran2025v} to evaluate the quality of video representations. Specifically, it consists of a single learnable query token cross-attending to the video tokens through a 12-head cross-attention block, followed by an MLP and a linear classifier for 15-way motion primitive classification.

\textbf{Attentive Probe Training and Evaluation.}
We train the attentive probe using 300 demonstrations per task used for policy training on RoboCasa Kitchen \citep{nasiriany2024robocasa}, and evaluate it on 50 newly generated demonstrations per task. We extract the intermediate future-token representations from the 19th layer of the frozen video backbone and train the probe for 10 epochs with the AdamW optimizer using a batch size of 128, a learning rate of $10^{-3}$, and weight decay of 0.05. We report the best validation top-1 accuracy over the 10 epochs.

\subsection{CKNNA}
To measure the similarity between representation spaces, we use Centered Kernel Nearest-Neighbor Alignment (CKNNA) \cite{huh2024platonic}, a relaxed variant of
Centered Kernel Alignment (CKA) \cite{kornblith2019similarity}.
We compute CKNNA on 4,096 clips randomly sampled from the validation set of attentive probing analysis.
For each WAM checkpoint, we extract the intermediate future-token representations from the 19th layer of the video backbone and mean-pool the 192 future-frame tokens to obtain a clip-level representation.
We use representations extracted from clean future observations ($\tau=0.0$) by the Base WAM (\ie DiT4DiT) trained at 100K iterations as a fixed reference, and measure their alignment with representations obtained from noise ($\tau=1.0$) at each checkpoint of the Base WAM and \sname from 20K to 100K iterations.

\subsection{Generated Video Quality and One-Step Distillation}
We evaluate one-step video generation quality using Fr\'echet Video Distance (FVD) \cite{unterthiner2018towards} between 2,048 generated videos and 2,048 randomly sampled clips from the RoboCasa Kitchen training set~\citep{nasiriany2024robocasa}.
For DMD2~\citep{yin2024improved}, we train a one-step student in two stages. First, we fine-tune the Wan2.2-TI2V-5B model~\citep{wan2025wan} on RoboCasa Kitchen~\citep{nasiriany2024robocasa} for 100K iterations with a global batch size of 64. Second, we distill the fine-tuned model’s EMA teacher into a one-step student, following the NVIDIA FastGen implementation~\citep{fastgen2026}, for 5K iterations with a global batch size of 16, without classifier-free guidance~\citep{ho2022classifier} and with a GAN loss weight of 0.1. We then use this one-step student as a frozen video model and train only a separate action decoder for 100K iterations with a global batch size of 64.

\newpage

\section{Additional Analysis}\label{appendix:additional_analysis}

\textbf{Effect of Action Conditioning Tokens.}
We examine how the choice of video latent tokens for action conditioning affects the performance of our base WAM, DiT4DiT~\citep{ma2026dit4dit}. As shown in \Cref{tab:appendix_action_condition_token}, using only future tokens performs best, while using both current and future tokens performs comparably. Both settings outperform the current-only setting, suggesting that providing future information is useful for robot control. 

\textbf{Quality of Future Representations Generated from Noise.} We investigate whether future representations generated from random noise capture task-relevant information during WAM training. Specifically, we use attentive probing to compare representations from the video model of the base WAM (\ie DiT4DiT \citep{ma2026dit4dit}) under three input conditions: ground-truth future observations ($\tau=0$), random noise ($\tau=1$), and the current observation repeated for all future frames ($\tau=0$). As shown in \Cref{tab:appendix_future_input_probe}, representations generated from random noise achieve a probing accuracy of 68.1\%, outperforming those extracted from the repeated current observation (63.0\%), suggesting that the video model can generate representations containing useful information about future dynamics. However, the performance remains below that obtained using ground-truth future observations (75.8\%), indicating a gap between representations generated from noise and those extracted from actual future observations.

\begin{table}[h]
\centering
\captionsetup{position=top}
\captionsetup[subtable]{skip=2pt}
\caption{\textbf{Additional analyses of the base WAM.}
We examine (a) which video latent tokens to use for action conditioning and (b) how well future video representations encode future motion with different inputs. In (b), we use an attentive probe to classify each video into a predefined motion primitive class using the corresponding representations.
Both analyses use the base WAM (DiT4DiT) on RoboCasa Kitchen. All 5B models are trained for 100K iterations with a batch size of 64. The default setting in (a) is highlighted in \colorbox{baselinecolor}{gray}.}
\label{tab:appendix_token_and_probe}

\begin{subtable}[t]{0.43\linewidth}
    \centering\small
    \caption{Action conditioning tokens}
    \label{tab:appendix_action_condition_token}
    \begin{adjustbox}{max width=\linewidth}
    \begin{tabular}{ccc}
        \toprule
        Current & Future & SR (\%) \\
        \midrule
        \textcolor{SJViolet}{\cmark} & \textcolor{SJRed}{\xmark} & 58.3 \\
        \baseline{\textcolor{SJRed}{\xmark}} &
        \baseline{\textcolor{SJViolet}{\cmark}} &
        \baseline{\textbf{61.4}} \\
        \textcolor{SJViolet}{\cmark} & \textcolor{SJViolet}{\cmark} & 60.9 \\
        \bottomrule
    \end{tabular}
    \end{adjustbox}
\end{subtable}\hfill%
\begin{subtable}[t]{0.55\linewidth}
    \centering\small
    \caption{Attentive probing}
    \label{tab:appendix_future_input_probe}
    \begin{adjustbox}{max width=\linewidth}
    \begin{tabular}{lc}
        \toprule
        Future Input & Acc. (\%) \\
        \midrule
        Ground-truth Future ($\tau=0$) & \textbf{75.8} \\
        Random Noise ($\tau=1$) & 68.1 \\
        Current Observation ($\tau=0$) & 63.0 \\
        \bottomrule
    \end{tabular}
    \end{adjustbox}
\end{subtable}
\end{table}

\textbf{How Do Human Videos Contribute to Robot Action Generation?} We investigate whether temporal information in human videos helps the model generate robot actions beyond merely providing object and goal information. To this end, we compare FRED co-trained on robot demonstrations and human videos with a variant in which all future frames in each human-video training batch are replaced with the current observation, while preserving the observation history and language instructions. As shown in \Cref{tab:appendix_human_future_dynamics}, the variant that uses static human videos for human-robot co-training achieves 0\% task success and 0\% instruction-following rates on human-to-robot skill transfer tasks, with the robot remaining nearly stationary for over one minute. These results suggest that temporal information in human videos plays an important role in human-to-robot skill transfer. 

\begin{table}[h]
\centering
\caption{\textbf{Effect of human future dynamics on robot action generation.}
We report the average success rate (SR; \%), the instruction-following rate (IF; \%), and the average arm movement with standard deviation (deg/s), computed over the 7 joints of the right arm, on human-to-robot skill transfer tasks.}
\label{tab:appendix_human_future_dynamics}
\begin{adjustbox}{max width=0.95\linewidth}
\begin{tabular}{lccc}
\toprule
Method
& SR (\%)
& IF (\%)
& Arm Movement (deg/s) \\
\midrule
\textbf{FRED (Ours)}
& \textbf{50.0}
& \textbf{96.4}
& $\mathbf{3.54 \pm 1.25}$ \\
w/o Human Future Dynamics
& 0.0
& 0.0
& $0.43 \pm 0.18$ \\
\bottomrule
\end{tabular}
\end{adjustbox}
\end{table}

\newpage

\section{Additional Simulation Benchmark Results} \label{appendix:additional_simul_results}

To further verify the effectiveness of our method, we additionally compare \sname with state-of-the-art robot foundation models on the RoboCasa Kitchen~\citep{nasiriany2024robocasa} benchmark. As shown in \Cref{tab:appendix_sim_results}, we find that \sname achieves the highest average success rate, demonstrating that our method not only exhibits strong generalization and transferability in real-world manipulation scenarios but also has the potential to achieve leading simulation benchmark performance when sufficient computational resources are available. We note that these results differ from those in \Cref{tab:sim_results} because we adjust the training configurations to be comparable to those of competing VLAs and WAMs. Specifically, we train \sname with a global batch size of 256 for 100K iterations, keeping all other hyperparameters unchanged, while the reported baseline results are taken from their original papers or official code repositories.

\begin{table}[h]
\centering
\caption{\textbf{Results on the RoboCasa Kitchen benchmark.}
We report the average success rate (\%) on RoboCasa Kitchen.}
\label{tab:appendix_sim_results}
\begin{adjustbox}{max width=0.95\linewidth}
\begin{tabular}{lc}
\toprule
Method & SR (\%) \\
\midrule
RLDX-1 \citep{kim2026rldx} & 70.8  \\
GR00T N1.7 \citep{nvidia2026gr00tn17} & 70.8   \\
Xiaomi-Robotics-1 \citep{team2026xiaomi} & 74.5 \\
\midrule
Video Policy \citep{liang2025video} & 66.0 \\
Cosmos Policy \citep{kim2026cosmos} & 67.1 \\
MetaWAM \citep{liang2026video} & 68.9 \\
GAM \citep{han2026geometric} & 69.4 \\
\textbf{\sname{} (Ours)}
& \textbf{75.4} \\
\bottomrule
\end{tabular}
\end{adjustbox}
\end{table}

\section{Flow Matching Process in \sname}
\label{appendix:prelim}

We briefly describe conditional flow matching \citep{lipman2022flow, albergo2025stochastic} as a common formulation for video and action generation.

Let $\mathbf{x}$ denote a clean target (\eg future video latents or an action sequence) and $\mathbf{c}$ denote the conditioning inputs (\eg task instructions) that define the conditional target distribution $p(\mathbf{x}|\mathbf{c})$. We construct a noisy input $\mathbf{x}^\tau$ by linearly interpolating between $\mathbf{x}$ and Gaussian noise $\boldsymbol{\epsilon} \sim \mathcal{N}(\mathbf{0},\mathbf{I})$:
\begin{equation}
    \mathbf{x}^\tau = (1-\tau)\mathbf{x} + \tau\boldsymbol{\epsilon},
    \quad \tau \in [0,1],
\end{equation}
where $\tau$ denotes the diffusion timestep, with $\tau=0$ corresponding to clean data and $\tau=1$ to pure noise. Our video and action generation models are each parameterized by a Diffusion Transformer (DiT) \cite{peebles2023scalable}, which serves as a conditional velocity network $\mathbf{v}_{\theta}(\mathbf{x}^\tau,\tau \mid \mathbf{c})$ trained to reconstruct the target velocity $\boldsymbol{\epsilon}-\mathbf{x}$:
\begin{equation}
    \mathcal{L}_{\mathrm{FM}}(\theta)
    =
\mathbb{E}_{(\mathbf{x},\mathbf{c}),\boldsymbol{\epsilon},\tau}
    \left[
        \left\|
            \mathbf{v}_{\theta}(\mathbf{x}^\tau,\tau | \mathbf{c})
            -(\boldsymbol{\epsilon}-\mathbf{x})
        \right\|_2^2
    \right].
\end{equation}
At inference, samples are generated from Gaussian noise by integrating the learned velocity from $\tau=1$ to  $\tau=0$. An Euler sampler \citep{lipman2022flow} performs this integration through repeated updates,
\begin{equation}
    \mathbf{x}^{\tau-\Delta\tau}
    =
    \mathbf{x}^\tau
    -\Delta\tau\,\mathbf{v}_{\theta}(\mathbf{x}^\tau,\tau | \mathbf{c}),
\end{equation}
where $\Delta\tau>0$ denotes the step size. In our one-step generation setting, we perform a single Euler step with $\Delta\tau=1$, \ie the velocity prediction $\mathbf{v}_{\theta}(\mathbf{x}^\tau,\tau \mid \mathbf{c})$ is computed from a single forward pass with $\mathbf{x}^{\tau=1}=\boldsymbol{\epsilon}$. During this forward pass, the velocity network's intermediate hidden states $\mathbf{h}$ are extracted to condition action generation.

\newpage

\section{Limitations and Future Work}
\label{appendix:limitations}

A potential limitation of \sname{} is its reliance on the language understanding capabilities of the underlying video diffusion backbone, as our training objective focuses on future representation learning and does not provide additional supervision for language grounding.
As shown in \Cref{tab:openarm_real_h2r}, without human-video co-training, \sname{} achieves 0\% instruction-following performance on the human skill transfer tasks, while pre-trained VLA baselines retain some instruction-following ability. Co-training with human videos substantially improves this performance, suggesting that stronger language grounding may further benefit generalization to unseen instructions.
Future work could combine the spatiotemporal priors of video diffusion models with the stronger language grounding of pre-trained vision-language models.

Another limitation is that human-to-robot transfer in \sname remains implicit: human and robot videos jointly train the video model, but we do not explicitly learn correspondences between their representations across embodiments.
Although human-video co-training substantially improves instruction following by enabling the model to learn meaningful temporal information from human videos (as shown in \Cref{tab:appendix_human_future_dynamics}), task success remains moderate. 
We encourage future work to explore explicit cross-embodiment representation learning to better transfer human-demonstrated skills to robot control.

As future work, we also believe it would be interesting to extend our approach to more scalable video diffusion models with stronger robot-domain priors.
Current pre-trained video diffusion models are trained predominantly on human videos, often requiring substantial adaptation for downstream robotic tasks.
Larger robot-pre-trained video backbones may provide a better starting point for future representation learning, consistent with the larger gains observed at the 14B scale in \Cref{tab:droid_real}.

\section{Full Benchmark Results for Real-world Experiments} \label{appendix:full_results}

\subsection{DROID}\label{appendix:full_results:droid}
We report the full per-task results of the DROID experiments in \Cref{tab:droid_appendix_progress}.

\begin{table*}[h]
\caption{\textbf{Per-task results on zero-shot environment generalization (DROID).}
Performance of policies trained on the DROID dataset without environment-specific fine-tuning. We report the average task progress rate (\%) for the in-domain long-horizon task, and the average success rate (\%) for other task types. For in-domain easy tasks, Env. 1 and Env. 2 denote two environments with different tabletop appearances. All other tasks are evaluated in Env. 1.
}
\label{tab:droid_appendix_progress}
\centering\small
\begin{adjustbox}{max width=\textwidth}
\begin{tabular}{l c ccc ccc c ccc cc}
\toprule
&
& \multicolumn{3}{c}{In-domain easy (Env. 1)}
& \multicolumn{3}{c}{In-domain easy (Env. 2)}
& \multicolumn{1}{c}{Long-horizon}
& \multicolumn{3}{c}{Out-of-domain}
& \multicolumn{2}{c}{Object grounding} \\
\cmidrule(lr){3-5}
\cmidrule(lr){6-8}
\cmidrule(lr){9-9}
\cmidrule(lr){10-12}
\cmidrule(lr){13-14}
Method & Backbone
& Banana & Microwave & Cube
& Banana & Microwave & Cube
& Clear Table
& Perch & Reveal & Stamp
& Cup & Cutlery \\
\midrule
$\pi_{0.5}$~\citep{intelligence2025pi_}
& --
& 100.0 & 58.3 & 8.3
& 83.3 & 41.7 & 16.7
& 53.3
& 0.0 & 8.3 & 8.3
& 25.0 & 8.3 \\

GR00T N1.7~\citep{nvidia2026gr00tn17}
& --
& 91.7 & 25.0 & 8.3
& 100.0 & 16.7 & 8.3
& 60.0
& 0.0 & 0.0 & 8.3
& 25.0 & 16.7 \\
\midrule

Fast-WAM~\citep{yuan2026fast}
& Wan2.2-5B
& 41.7 & 33.3 & 0.0
& 50.0 & 41.7 & 0.0
& 26.7
& 0.0 & 0.0 & 0.0
& 8.3 & 8.3 \\

DiT4DiT~\citep{ma2026dit4dit}
& Wan2.2-5B
& 33.3 & 8.3 & 8.3
& 16.7 & 8.3 & 0.0
& 16.7
& 0.0 & 0.0 & 0.0
& 8.3 & 8.3 \\

AGRA~\citep{qiu2026making}
& Wan2.2-5B
& 100.0 & 25.0 & 0.0
& 83.3 & 33.3 & 16.7
& 31.7
& 0.0 & 8.3 & 8.3
& 25.0 & 41.7 \\

\textbf{\sname{} (Ours)}
& Wan2.2-5B
& 83.3 & 25.0 & 16.7
& 75.0 & 33.3 & 50.0
& 46.7
& 0.0 & 41.7 & 16.7
& 41.7 & 33.3 \\
\midrule

DreamZero~\citep{ye2026world}
& Wan2.1-14B
& 83.3 & 50.0 & 58.3
& 91.7 & 41.7 & 41.7
& 61.7
& 16.7 & 0.0 & 16.7
& 8.3 & 25.0 \\

\textbf{\sname{} (Ours)}
& Wan2.1-14B
& 91.7 & 66.7 & 75.0
& 83.3 & 58.3 & 41.7
& 81.7
& 33.3 & 33.3 & 50.0
& 50.0 & 58.3 \\
\bottomrule
\end{tabular}
\end{adjustbox}
\end{table*}

\newpage
\subsection{OpenArm}\label{appendix:full_results:openarm}

We report the full per-task results of the OpenArm experiments in \Cref{tab:openarm_appendix_progress}.

\begin{table}[H]
\centering
\caption{\textbf{Per-task results on humanoid control with and without human-robot co-training.}
Performance of policies trained on the OpenArm humanoid using robot demonstrations only or with additional human videos.
We report average success rate (\%) on in-domain atomic tasks, average task progress rate (\%) on the in-domain composite task, and both task success (\%) and instruction-following rate (\%) on human skill transfer tasks.}
\label{tab:openarm_appendix_progress}
\vspace{-0.5em}
\small
\begin{subtable}[t]{\textwidth}
\centering
\caption{Results on robot in-domain tasks.}
\label{tab:openarm_appendix_indomain}
\begin{adjustbox}{max width=0.8\textwidth}

\begin{tabular}{l c ccc c}
\toprule
&
&
\multicolumn{3}{c}{Atomic}
&
\multicolumn{1}{c}{Composite}
\\
\cmidrule(lr){3-5}
\cmidrule(lr){6-6}
Method
& Training Data
& PnP Banana
& \makecell{Air Fryer}
& \makecell{PnP Apple/Mango}
& \makecell{Air Fryer + Apple}
\\
\midrule
$\pi_{0.5}$~\citep{intelligence2025pi_}
& --
& 25.0 & 100.0 & 66.7 & 50.0
\\
GR00T N1.7~\citep{nvidia2026gr00tn17}
& --
& 91.7 & 100.0 & 83.3 & 45.8
\\
\midrule
DreamZero~\citep{ye2026world}
& Robot only
& 50.0 & 100.0 & 33.3 & 0.0
\\
Fast-WAM~\citep{yuan2026fast}
& Robot only
& 25.0 & 66.7 & 75.0 & 0.0
\\
DiT4DiT~\citep{ma2026dit4dit}
& Robot only
& 83.3 & 100.0 & 50.0 & 50.0
\\
AGRA~\citep{qiu2026making}
& Robot only
& 75.0 & 91.7 & 75.0 & 8.3
\\
\textbf{\sname{} (Ours)}
& Robot only
& 100.0 & 100.0 & 75.0 & 70.8
\\
\midrule
DreamZero~\citep{ye2026world}
& Robot + human
& 66.7 & 100.0 & 8.3 & 0.0
\\
Fast-WAM~\citep{yuan2026fast}
& Robot + human
& 25.0 & 91.7 & 75.0 & 0.0
\\
DiT4DiT~\citep{ma2026dit4dit}
& Robot + human
& 83.3 & 100.0 & 75.0 & 41.7
\\
AGRA~\citep{qiu2026making}
& Robot + human
& 50.0 & 83.3 & 91.7 & 12.5
\\
\textbf{\sname{} (Ours)}
& Robot + human
& 100.0 & 100.0 & 91.7 & 75.0
\\
\bottomrule
\end{tabular}
\end{adjustbox}
\end{subtable}

\vspace{0.6em}
\begin{subtable}[t]{\textwidth}
\centering
\caption{Task success rates (\%) on human skill transfer tasks.}
\label{tab:openarm_h2r_success}
\begin{adjustbox}{max width=1.0\textwidth}
\begin{tabular}{l c cc cccc c}
\toprule
Method & Training Data
& Bottle on Basket
& Box to Center 
& PnP Bottle & PnP Box & PnP Ball & PnP Doll
& Pour Coffee \\
\midrule
$\pi_{0.5}$~\citep{intelligence2025pi_}
& -- & 0.0 & 0.0 & 0.0 & 0.0 & 0.0 & 0.0 & 0.0 \\
GR00T N1.7~\citep{nvidia2026gr00tn17}
& -- & 0.0 & 0.0 & 8.3 & 0.0 & 16.7 & 41.7 & 0.0 \\
\midrule
DreamZero~\citep{ye2026world}
& Robot only & 0.0 & 0.0 & 0.0 & 0.0 & 0.0 & 0.0 & 0.0 \\
Fast-WAM~\citep{yuan2026fast}
& Robot only & 0.0 & 0.0 & 0.0 & 0.0 & 0.0 & 0.0 & 0.0 \\
DiT4DiT~\citep{ma2026dit4dit}
& Robot only & 0.0 & 0.0 & 0.0 & 0.0 & 0.0 & 0.0 & 0.0 \\
AGRA~\citep{qiu2026making}
& Robot only & 0.0 & 0.0 & 0.0 & 0.0 & 0.0 & 0.0 & 0.0 \\
\textbf{\sname{} (Ours)}
& Robot only & 0.0 & 0.0 & 0.0 & 0.0 & 0.0 & 0.0 & 0.0 \\
\midrule
DreamZero~\citep{ye2026world}
& Robot + human & 0.0 & 0.0 & 8.3 & 0.0 & 0.0 & 0.0 & 0.0 \\
Fast-WAM~\citep{yuan2026fast}
& Robot + human & 0.0 & 0.0 & 0.0 & 0.0 & 0.0 & 0.0 & 0.0 \\
DiT4DiT~\citep{ma2026dit4dit}
& Robot + human & 0.0 & 0.0 & 0.0 & 25.0 & 0.0 & 25.0 & 25.0 \\
AGRA~\citep{qiu2026making}
& Robot + human & 0.0 & 0.0 & 0.0 & 0.0 & 0.0 & 0.0 & 0.0 \\
\textbf{\sname{} (Ours)}
& Robot + human & 25.0 & 33.3 & 66.7 & 50.0 & 41.7 & 75.0 & 58.3 \\
\bottomrule
\end{tabular}
\end{adjustbox}
\end{subtable}

\vspace{0.8em}

\begin{subtable}[t]{\textwidth}
\centering
\caption{Instruction-following rates (\%) on human skill transfer tasks.}
\label{tab:openarm_h2r_if}
\begin{adjustbox}{max width=1.0\textwidth}
\begin{tabular}{l c cc cccc c}
\toprule
Method & Training Data
& Bottle on Basket
& Box to Center 
& PnP Bottle & PnP Box & PnP Ball & PnP Doll
& Pour Coffee \\
\midrule
$\pi_{0.5}$~\citep{intelligence2025pi_}
& -- & 25.0 & 75.0 & 0.0 & 0.0 & 0.0 & 0.0 & 25.0 \\
GR00T N1.7~\citep{nvidia2026gr00tn17}
& -- & 0.0 & 66.7 & 100.0 & 100.0 & 100.0 & 100.0 & 0.0 \\
\midrule
DreamZero~\citep{ye2026world}
& Robot only & 0.0 & 0.0 & 0.0 & 0.0 & 0.0 & 0.0 & 0.0 \\
Fast-WAM~\citep{yuan2026fast}
& Robot only & 0.0 & 0.0 & 0.0 & 0.0 & 0.0 & 0.0 & 0.0 \\
DiT4DiT~\citep{ma2026dit4dit}
& Robot only & 0.0 & 0.0 & 0.0 & 0.0 & 0.0 & 0.0 & 0.0 \\
AGRA~\citep{qiu2026making}
& Robot only & 0.0 & 0.0 & 0.0 & 0.0 & 0.0 & 0.0 & 0.0 \\
\textbf{\sname{} (Ours)}
& Robot only & 0.0 & 0.0 & 0.0 & 0.0 & 0.0 & 0.0 & 0.0 \\
\midrule
DreamZero~\citep{ye2026world}
& Robot + human & 33.3 & 0.0 & 25.0 & 0.0 & 0.0 & 8.3 & 0.0 \\
Fast-WAM~\citep{yuan2026fast}
& Robot + human & 16.7 & 25.0 & 0.0 & 0.0 & 0.0 & 0.0 & 16.7 \\
DiT4DiT~\citep{ma2026dit4dit}
& Robot + human & 0.0 & 8.3 & 33.3 & 58.3 & 50.0 & 58.3 & 91.7 \\
AGRA~\citep{qiu2026making}
& Robot + human & 0.0 & 0.0 & 0.0 & 0.0 & 0.0 & 0.0 & 0.0 \\
\textbf{\sname{} (Ours)}
& Robot + human & 91.7 & 83.3 & 100.0 & 100.0 & 100.0 & 100.0 & 100.0 \\
\bottomrule
\end{tabular}
\end{adjustbox}
\end{subtable}
\end{table}

\end{document}